\documentclass{article}
\usepackage[T1]{fontenc}
\usepackage[utf8]{inputenc}
\usepackage{amsmath,amsfonts,amsthm,amssymb}
\usepackage{setspace}
\usepackage{fullpage}
\usepackage{fancyhdr}
\usepackage{xspace, xcolor}

\usepackage{verbatim}
\usepackage{lastpage}
\usepackage{extramarks}
\usepackage{natbib}
\usepackage{diagbox}
\usepackage{booktabs}
\usepackage{titletoc}

\usepackage{times}
\usepackage{setspace}
\usepackage{graphicx,float,wrapfig}
\usepackage{algorithm}
\usepackage{enumitem}
\usepackage[noend]{algpseudocode}
\usepackage{subcaption}
\usepackage{xspace}
\usepackage{multirow}

\usepackage{xltabular}
\newcolumntype{L}{>{\RaggedRight\hspace{0pt}}X}

\allowdisplaybreaks

\theoremstyle{definition}

\def\shownotes{1}  %
\ifnum\shownotes=1
\newcommand{\authnote}[2]{[#1: #2]}
\else
\newcommand{\authnote}[2]{}
\fi

\usepackage{hyperref}
\hypersetup{
	colorlinks   = true, %
	urlcolor     = red, %
	linkcolor    = blue, %
	citecolor   = blue %
}

\usepackage{amsmath,amsfonts,bm}

\newcommand*{\defeq}{\triangleq}

\def\1{\bm{1}}

\makeatletter
\newcommand{\ve}{\@ifnextchar\bgroup{\velong}{{\bm{e}}}}
\newcommand{\velong}[1]{{\bm{#1}}}
\makeatother

\DeclareMathAlphabet{\mathsfit}{\encodingdefault}{\sfdefault}{m}{sl}
\SetMathAlphabet{\mathsfit}{bold}{\encodingdefault}{\sfdefault}{bx}{n}

\def\calD{{\mathcal{D}}}

\def\calX{{\mathcal{X}}}

\DeclareMathOperator*{\argmin}{argmin}

\def\({\left(}
\def\){\right)}
\def\[{\left[}
\def\]{\right]}

\DeclareMathOperator{\supp}{supp}
\newcommand{\ind}[1]{\mathbb{I}\left[ #1 \right]}

\newcommand{\algname}{STP\xspace}

\newcommand{\len}{\text{len}}

\definecolor{keywordcolor}{rgb}{0.7, 0.1, 0.1}   %
\definecolor{tacticcolor}{rgb}{0.0, 0.1, 0.6}    %
\definecolor{commentcolor}{rgb}{0.4, 0.4, 0.4}   %
\definecolor{symbolcolor}{rgb}{0.0, 0.1, 0.6}    %
\definecolor{sortcolor}{rgb}{0.1, 0.5, 0.1}      %
\definecolor{attributecolor}{rgb}{0.7, 0.1, 0.1} %
\usepackage{listings}

\begin{document}

\title{STP: Self-play LLM Theorem Provers with Iterative \\Conjecturing and Proving}

\author{Kefan Dong \\ 
	Stanford University \\
	\texttt{kefandong@stanford.edu}
	\and
	Tengyu Ma \\
	Stanford University \\
	\texttt{tengyuma@stanford.edu}
}
\date{}

\maketitle
\begin{abstract}
\noindent
A fundamental challenge in formal theorem proving by LLMs is the lack of high-quality training data. 
Although reinforcement learning or expert iteration partially mitigates this issue by alternating between LLM generating  proofs and finetuning them on correctly generated ones, performance quickly plateaus due to the scarcity of correct proofs (sparse rewards).
To keep improving the models with limited data, we draw inspiration from mathematicians, who continuously develop new results, partly by proposing novel conjectures or exercises (which are often variants of known results) and attempting to solve them.
We design the Self-play Theorem Prover (\algname) that simultaneously takes on two roles, conjecturer and prover, each providing training signals to the other. 
The conjecturer is trained iteratively on previously generated conjectures that are barely provable by the current prover, which incentivizes it to generate increasingly challenging conjectures over time. The prover attempts to prove the conjectures with standard expert iteration. 
We evaluate \algname with both Lean and Isabelle formal versifiers. 
With 51.3 billion tokens generated during the training in Lean, \algname proves 28.5\% of the statements in the LeanWorkbook dataset,  doubling the previous best result of 13.2\% achieved through expert iteration.
The final model achieves state-of-the-art performance among whole-proof generation methods on miniF2F-test (65.0\%, pass@3200), ProofNet-test (23.9\%, pass@3200) and PutnamBench (8/644, pass@3200). We release our code, model, and dataset in this url: \url{https://github.com/kfdong/STP}.
\end{abstract}

\section{Introduction}
\begin{figure*}[htp]
	\centering
	\includegraphics[width=0.9\linewidth]{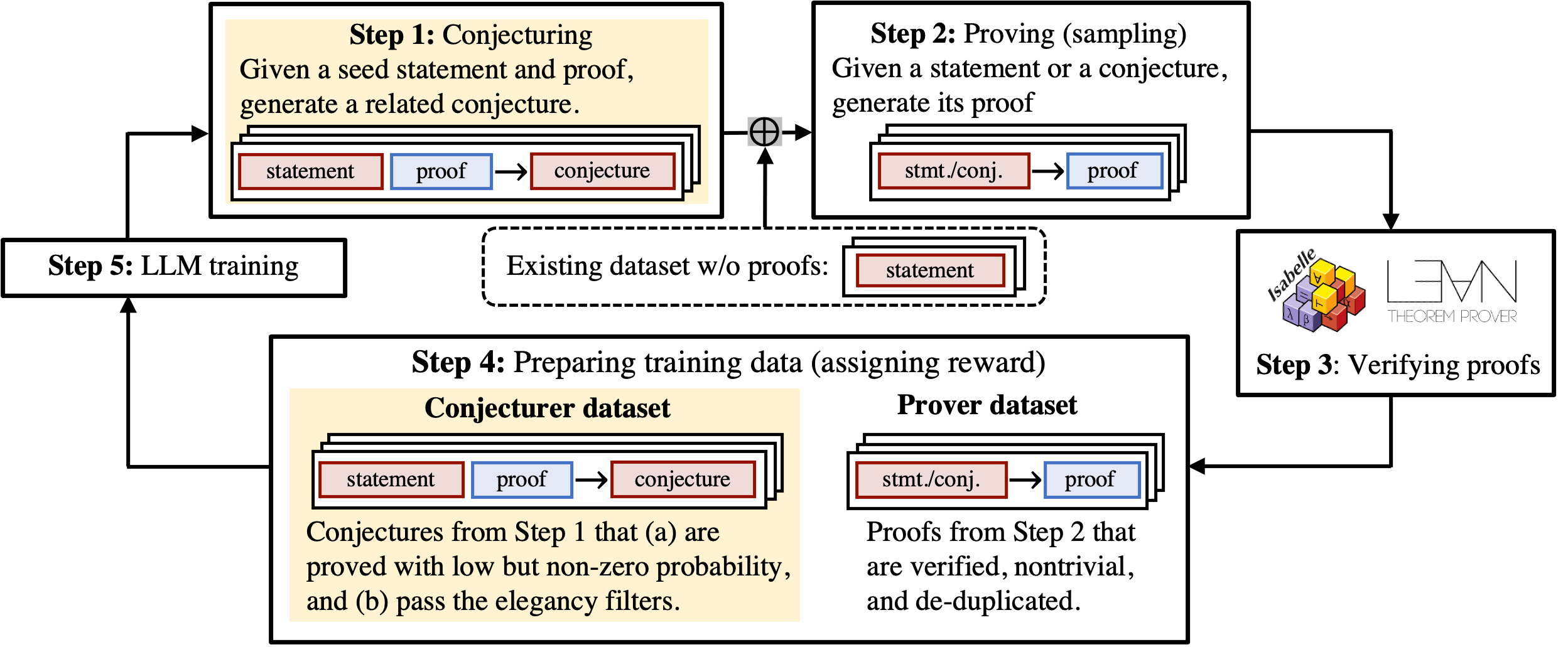}
	\caption{
		Self-play Theorem Prover (\algname). 
		Our model simultaneously takes on two roles --- the conjecturer that generates new, related conjecture given a seed theorem with proof (Step 1), and the prover that attempts to prove the statements in an existing dataset and the generated conjectures (Step 2). Step 4 selects the correct, approachable, elegant, yet challenging conjectures to train the conjecturer, and the verifier selects correct proofs in Step 3 to train the prover.
		The main difference between \algname and expert iteration is the conjecturer role highlighted with a yellow background.
		}
	\label{fig:main}
\end{figure*}

The reasoning capability of large language models (LLMs) is critical for various applications, including coding assistants, question-answering, and agents \citep{plaat2024reasoning,shinn2023reflexion,yao2022react,shao2024deepseekmath,li2023starcoder,nijkamp2022codegen}. It is also a key criterion for achieving artificial general intelligence (AGI). 
Automated theorem proving with formal languages by LLMs stands at the forefront of reasoning research \citep{yang2024formal}, partly because it allows objective and reliable evaluation through classical verifiers such as Lean \citep{moura2021lean} and Isabelle \citep{nipkow2002isabelle}. Moreover, it arguably encapsulates the essence of advanced reasoning tasks while abstracting away the ambiguity of natural language, enabling meaningful studies on a relatively smaller scale.

\sloppy However, a fundamental challenge in improving reasoning performance—whether in natural or formal languages—lies in the lack of high-quality training data. Collecting reasoning data requires domain experts, making it expensive to scale.
There are only a limited number of advanced math papers and theorems in existence, orders of magnitude smaller than other data sources.

Reinforcement learning (RL) on datasets \textit{without} solutions (e.g., datasets with theorem statements or reasoning questions and answers) is a prominent approach for improving the reasoning capability, as seen in the recent development of OpenAI o1~\cite{jaech2024openai}, DeepSeek-Prover~\citep{xin2024deepseeka} and DeepSeek R1~\citep{guo2025deepseek}. Often referred to as expert iteration~\citep{anthony2017thinking}, it partially mitigates the data scarcity issue by alternating between LLMs generating proofs and finetuning them on correctly generated ones \citep{kaliszyk2018reinforcement, wu2021tacticzero, alphaproof2024, xin2024deepseek, ying2024lean}. 

However, as \citet{wu2024internlm2} pointed out, RL or expert iteration often saturates at a low pass rate because the number of samples required to generate a correct proof for an unproven theorem grows exponentially. As a result, a massive amount of computation is wasted on generating incorrect proofs that provide no training signal to the model. For instance, in the proof sampling process of \citet{wu2024internlm2}, 98.5\% of the compute yields no successful proofs, despite the pass rate being only 13.2\% on the training dataset, LeanWorkbook \citep{ying2024lean}. In other words, after a few rounds of expert iteration, re-training the model becomes much less effective due to the limited number of new successful proofs.

In addition, RL’s capability is fundamentally bounded by the difficulty level of the theorems in the training dataset—it is unlikely, in principle, for a model to learn college-level proof techniques solely by working on high school-level problems or to solve open math problems using RL on graduate-level problems. Moreover, there are likely not enough open problem statements available for RL training to generalize to other open problems, particularly more advanced ones. In other words, RL or expert iteration algorithms cannot train indefinitely without continuously collecting more theorem statements or math problems.

We need an algorithm that can run and self-improve indefinitely \emph{without more data}. To this end, we draw inspiration from how mathematicians learn and develop advanced mathematics; they refine their understanding and sharpen their proof skills by working on synthesized exercises—variants, extensions, or combinations of existing theorems. Additionally, they frequently propose and publish conjectures, a process widely regarded as just as important, if not more so, than solving them.
In other words, unlike the current training of LLMs, mathematicians engage with far more exercises and conjectures (referred to collectively as conjectures in this paper) than the polished, published results found in academic papers and books. Moreover, the continuous generation of new conjectures keeps mathematical fields dynamic and moving forward.

In this paper, we design Self-play Theorem Prover (\algname), which mimics how mathematicians learn and develop mathematics. It simultaneously assumes two roles—conjecturer and prover—providing training signals to each other.

As illustrated in Fig.~\ref{fig:main}, the conjecturer, given a seed theorem with proof, proposes a new, related conjecture (Step 1), while the prover attempts to prove conjectures and statements from an existing dataset (Step 2). Then, the verifier selects correct proofs (Step 3) to train the prover using standard RL and identifies correct, approachable, elegant, yet challenging conjectures to supervise the training of the conjecturer (Step 4).
More concretely, in each iteration, the conjecturer is trained on previously generated conjectures that:
(a) are barely provable by the current prover (i.e., the prover’s success probability with respect to its random seed is positive but low), and
(b) pass certain elegancy filters.
This iterative process gradually increases the difficulty of conjectures and proofs without requiring additional  data. Our method can be viewed either as a self-play algorithm between conjectures and provers or as automated curriculum learning \citep{portelas2020automatic} with a self-generated adaptive curriculum (via conjecturers).

\begin{figure}[tp]
	\centering
	\begin{minipage}[t]{0.46\textwidth}
		\centering
		\includegraphics[width=0.965\linewidth]{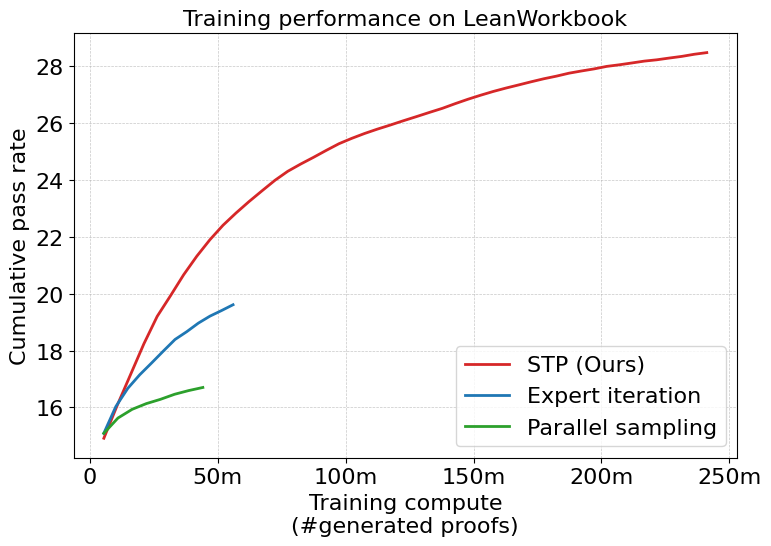}
		\caption{The cumulative pass rates of \algname, expert iteration, and parallel sampling on LeanWorkbook  shows that \algname achieves a much better scaling in terms of the performance vs number of generated proofs. The compute for generating conjectures and training the conjecturer in \algname is negligible because the number of generated proofs during training is 64 times the number of conjectures.}
		\label{fig:lean-main}
	\end{minipage}
	~~~~~~~
	 \raisebox{-0.9pt}{%
		\begin{minipage}[t]{0.46\textwidth}
			\centering
			\vspace*{-5.25cm}\includegraphics[width=\linewidth]{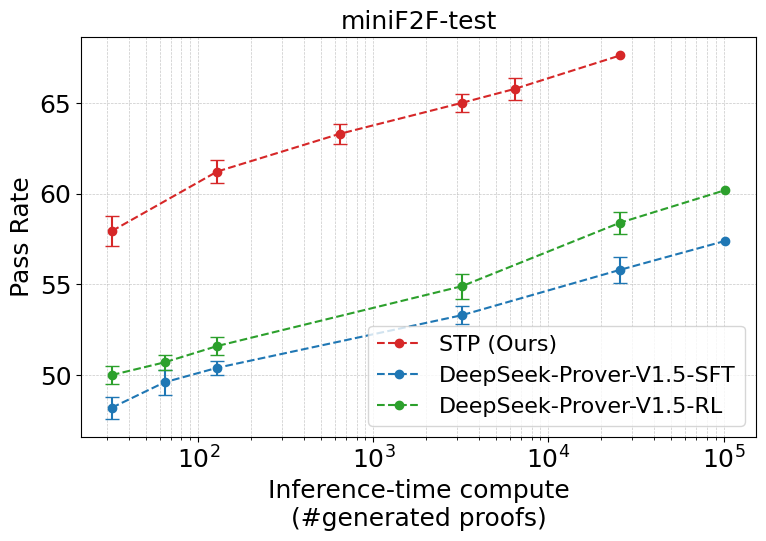}
			\caption{Comparison of pass rates on miniF2F-test (y-axis) with different numbers of inference-time samples (x-axis). The model trained with \algname consistently outperforms the DeepSeek-Prover-V1.5 series.}
			\label{fig:scatter-main}
		\end{minipage}%
	}
\end{figure}

We empirically evaluate our method with both Lean~\citep{moura2021lean} and Isabelle~\citep{nipkow2002isabelle}. For the Lean experiments, we aim for the best performance and therefore choose DeepSeek-Prover-V1.5-SFT~\citep{xin2024deepseek} as the base model for \algname. As shown in Fig.~\ref{fig:lean-main}, after a self-play training of roughly 241M generated proofs and 3.6M generated conjectures, we successfully prove 28.5\% of the statements in the training dataset LeanWorkbook~\citep{ying2024lean},  doubling the previous best result of 13.2\% \citep{wu2024internlm2} achieved by expert iteration.
In Fig.~\ref{fig:scatter-main}, we compare the inference-time performance of existing models and the final model trained with \algname by taking multiple independent samples on a common benchmark, miniF2F-test \citep{zheng2021minif2f}.
Our model significantly outperforms the DeepSeek-Prover-V1.5 models across various sampling budgets. We also achieve state-of-the-art performance among whole-proof generation methods on miniF2F-test (65.0\%, pass@3200), ProofNet-test (23.9\%, pass@3200) \citep{azerbayev2023proofnet} and PutnamBench (8/644, pass@3200) \citep{tsoukalas2024putnambench}, where pass@k represents the percentage of statements proved with $k$ independently sampled proofs per statement.

In the Isabelle experiments, we study the scalability of \algname by starting from a generic math-focused model Llemma-7b~\citep{azerbayev2023llemma} and run \algname for more iterations (300M generated proofs in total). We compare the scaling of \algname with expert iteration and parallel sampling, by taking several model checkpoints during the \algname training run and then switching to the baseline methods. The results clearly demonstrate that \algname achieves a better scaling behavior starting from various checkpoints with different capability (see Fig.~\ref{fig:isabelle-main} (Left) in Section~\ref{sec:results-isabelle}). 
Ablation study also demonstrates that the main performance gain stems from the dense training signals given by the conjectures. Expert iteration wasted its compute on generating unsuccessful proofs to challenging theorems in the training dataset---at a checkpoint where the pass rate is around 11.4\% on LeanWorkbook, only 131 out of 2.5M generated proofs of the unproved statements are correct, resulting in very limited training signals. In contrast, at least 47\% of the generated conjectures in STP training are successfully proved because the conjecturer is trained to generate more approachable statements thanks to the design of its reward (see Fig.~\ref{fig:isabelle-main} (Right)).

\section{Additional Related Works}\label{sec:related-works}

We refer the readers to \citet{bibel2013automated,loveland2016automated} and the reference therein for classical automated theorem proving. Below, we discuss recent works on modern LLM-based theorem provers in addition to what has been discussed in the intro. 

\paragraph{Autoformalization.} A relatively efficient way to create formal proof data is autoformalization, that is, translating natural language math statements and/or proofs to formal language \citep{jiang2023multilingual, lu2024process}. A line of research focuses on generating proofs or reasoning steps in natural language and then formalizing the proofs \citep{jiang2022draft, zheng2023lyra, wang2023lego}. Most recently, \citet{alphaproof2024, xin2024deepseeka, xin2024deepseek} autoformalize statements and then train with expert iteration / RL to write proofs, achieving significant improvement over prior works thanks to the large-scale natural language datasets.

\paragraph{Formal conjecturing.} Prior works also study how to generate new formal statements/conjectures by neural networks \citep{urban2020first, einarsdottir2024lemma, johansson2023exploring} or human-written generators \citep{polu2022formal, trinh2024alphageometry}, and find that the synthetic statements are generally useful for training the provers \citep{wang2020learning, wu2020int}. Synthetic statements and proofs can also be extracted from an incorrect proof trajectory during RL with hindsight experience replay (HER)~\cite{andrychowicz2017hindsight} to speed up the training process \citep{aygun2022proving, dong2024formal}. However, even though the training efficiency is improved, we argue that the final performance is still bounded by difficulty level of the existing dataset because synthetic statements are most likely easier than the given ones in the dataset.

\paragraph{Self-play and automatic goal generation.} The closest related work to this paper is \citet{poesia2024learning} which also designs a self-play training that iterates between conjecturing and theorem proving. The key difference between this paper and \citet{poesia2024learning} is that we start with a pre-trained model and work on practical formal languages like Lean and Isabelle with an infinite space of possible proof steps (which are actions in the RL algorithm), whereas \citet{poesia2024learning} operates in a simplified and constrained setting with a finite action space and trains from scratch. As a result, \citet{poesia2024learning} rely on constrained decoding to force the validity of generated conjectures, while we solely rely on the LLM itself to generate valid conjectures. Technically, since our training process is much longer (more than 50 iterations) than \citet{poesia2024learning} (5 iterations), we must carefully design the conjecturing reward to maintain the diversity and relevance of the generated conjectures (see Section~\ref{sec:stp-loop}). 

The idea of generating new tasks by the model is also explored in other domains such as alignment \citep{ye2024evolving}, programming puzzles \citep{haluptzok2022language, teodorescu2023codeplay, pourcel2024aces}, video games \citep{zhang2023omni, pourcel2024autotelic}, and classic RL environments \citep{parker2022evolving, colas2022autotelic}.
More generally, self-play training has demonstrated its potential to achieve super-human performance on two-player games in a fixed environment like Go \citep{silver2017mastering}.\nocite{sun2018dual}

\section{Method}\label{sec:method}
On the high level, Self-play Theorem Prover (\algname) involves three training stages: (1) model initialization by supervised finetuning, (2) self-play training (visualized in Fig.~\ref{fig:main}), and (3) final re-training. Unless otherwise stated, we use the term `statement' to refer to the statements in given datasets, and `conjecture' the generated conjectures.

\subsection{Model initialization by supervised finetuning}\label{sec:sft}

In this stage, we initialize the model with two roles, conjecturer and prover, by finetuning a generic LLM (such as the Llama~\citep{touvron2023llama}) on a SFT dataset constructed from existing proof libraries such as Mathlib~\citep{mathlib}. The proof libraries are organized into files containing human-written formal proofs of known mathematical theorems, and each file formalizes a relatively self-contained result, such as a chapter of a textbook. Our SFT data consists of the following two parts, for finetuning the prover and conjecturer, respectively. Also see concrete examples in Appendix~\ref{app:examples}.

\paragraph{Prover SFT dataset.} We construct a SFT dataset to teach the model to write formal proofs in the given format, where each example is the concatenation of a system prompt (to instruct the model to generate in formal language), a statement, and its corresponding proof. We only compute the next token prediction loss on the proof (which is the expected output of the model), while the rest is treated as input. To build this dataset, we simply extract all the statement-proof pairs in the proof library files and add a system prompt.

\paragraph{Conjecturer SFT dataset.} Generally, the conjecturer is to generate a new, related conjecture, given a seed statement with proof that provide the initial ideas. Technically, to further guide the generation of conjecturer, we also provide it a lemma used in the proof of the seed statement, which can be extracted from the verifier,\footnote{There is no fundamental difference between lemmas and theorems in formal proofs --- the naming is purely for better exposition.} so that the generated conjectures are more likely to be related to the theorem through the lemma.  Therefore, the input  is a concatenation of the system prompt, a lemma, and a seed statement and its proof, separated by special formatting tokens, and the expected output  is a conjecture on which we compute training loss. We also allow the model to generate conjectures with a fixed trivial lemma. 
To construct this dataset, we extract (lemma, theorem X, theorem Y) tuples from every proof library file such that (a) the lemma and two theorems appears in the file in this particular order, and (b) the lemma is used in the proof of both theorems. The lemma and theorem X will be part of the inputs, and theorem Y will be the output.

\subsection{Self-play training}\label{sec:stp-loop} 
Our self-play training stage of \algname is shown in Fig.~\ref{fig:main}. The main difference compared to expert iteration is the conjecturer in Steps 1 and 4, highlighted in a yellow background.

\paragraph{Generating conjectures and proofs (Steps 1 \& 2).} The self-play training starts with collecting a list of the conjecturer's inputs in the same format as in the conjecture SFT dataset (system prompt, lemma, and theorem), but from theorem-proof pairs where the theorems are from the given dataset without proofs and proofs are previously generated. We extract a seed lemma from the proof, using the verifier.\footnote{In our implementation, lemmas are extracted together with proof verification in Step 3 by configuring the verifiers accordingly.} To prevent the model from only focusing on a few particular proof techniques, we de-duplicate the list based on the seed statement and lemma, and randomly drop some inputs whose lemma appears excessively.
Then, the LLM generates conjectures from the inputs, and we randomly select a subset of the generated conjectures with size no larger than the number of remaining unproved statements in the given dataset, so that the prover's compute budget is split equally between the conjectures and statements. (See the pseudo-code and details in Appendix~\ref{app:step1-pseudo-code}.)
For the prover's inputs, we combine the generated conjectures and the unproved statements in the existing dataset. Then, we independently sample $K$ proofs per statement/conjecture in Step 2.

\paragraph{Reward assignments (Step 4).} The major technical challenge of \algname is to design the reward function for the conjecturer (in other words, construct the conjecturer dataset in Step 4). The ultimate goal is to incentivize conjecturer to generate diverse, relevant, approachable yet challenging conjectures to provide enough training signals to the prover.

In Step 4,  we first organize all generated conjectures and proofs into a list of examples $\calD=\{(t_i,p^t_i,l_i,c_i,p^c_i)\}_{i=1}^n$ where $t_i$ and $p^t_i$ represents a seed statement and its proof, $l_i$ is a lemma used in the proof $p^t_i$, and $c_i,p^c_i$ are the generated conjectures and the generated proof.  We will filter $\calD$ as described below and then use $(t_i,p^t_i,l_i)$ as the input to the conjecturer and $c_i$ as the output, and $p^c_i$ as the output of the prover w.r.t. the input $c_i.$

To decide whether a conjecture $c$ is challenging, we use the (empirical) pass rate of the prover estimated by the $K$ independently generated proofs:
\begin{equation*}\textstyle{\hat{P}(c)\defeq (\#\{i:c_i=c,p^c_i\text{ is correct}\}) / (\#\{i:c_i=c\}).}\end{equation*}
Then, we select the examples in $\calD$ where (a) lemma $l_i$ is used in the proof of conjecture $p^c_i$, and (b) the pass rate of the conjecture, $\hat{P}(c_i)$, is between $(0,1/4]$:
\begin{align*}\overline{\calD}\gets \{&(t_i, p^t_i, l_i, c_i) \mid (t_i, p^t_i, l_i, c_i, p^c_i)\in \calD, \\&\quad \hat{P}(c_i)\in (0,1/4], p^c_i\text{ is correct}, l_i\text{ is used in }p^c_i\}.
	\end{align*}
Here we discard the proofs (of the conjecture) $p^c_i$ since they are not needed to train the conjecturer, and we remove the duplicated conjectures (that have multiple proofs).

Then, we apply a heuristic elegancy filter to discourage the model from generating artificially hard conjectures with complicated goals --- we remove conjectures whose minimum proof length divided by the length of the conjecture is in the lowest 20\% of remaining examples.

Finally, we re-weight the selected conjectures to preserve the diversity of the conjecturer --- the reward for conjecturer cannot only depend on the generated conjectures individually because otherwise the conjecturer's optimal policy may degenerate to a singular distribution, whereas in reality, the given dataset typically has multiple modes because the statements focus on different topics like algebra, number theory, and calculus. Therefore, our idea is to push the \emph{distribution} of the selected conjectures toward the unproved statements in the existing dataset to maintain the balance between multiple modes. 
To this end, we compute a distribution $P$ supported on the selected conjectures that minimizes the Wasserstein distance to the uniform distribution over unproved theorems, denoted by $Q$. 
The matching cost or similarity metric between a conjecture and a statement, used for computing the Wasserstein distance between $P$ and $Q$, is defined as the negative cosine similarity between their embeddings (given by the current model). Finally, we use the distribution $P$ as the training set for the conjecturer. 
Pseudo-code of this step is in Appendix~\ref{app:step4-pesudo-code}, and an efficient implementation is in Appendix~\ref{app:wasserstein}. 

For the prover dataset, we only select correct generated proofs where the empirical pass rate of the corresponding statement/conjecture is below 1/2. (We consider other correct proofs trivial).  We de-duplicate the prover dataset by exact match. Then, the prover is trained on a replay buffer containing the selected proofs from the last three iterations.

\paragraph{LLM training (Step 5).} We use weighted cross entropy loss computed on the conjectures or proofs (but not the inputs of the model). For the proof dataset, we weight the examples reciprocally to the number of verified proofs to the corresponding statement/conjecture. We also use a length penalization of the form $\gamma^{L}$ to reward simpler proofs, where $\gamma<1$ is the discount factor and $L$ is the length of the proof. For the experiments with Lean, we additionally reward proofs that has faster verification time by a penalization of the form $\beta^{T}$, where $T$ is the execution time of the Lean verifier.\footnote{In our preliminary experiments, we found that without the penalization on verification time, the Lean verifier takes 2x more wall-clock time on CPU than sampling proofs on TPU for our cluster setup, which becomes a bottleneck for STP training.}

\subsection{Final re-training}\label{sec:checkpointing}
To avoid training instability caused by the changing data distribution during self-play, we re-train the final model checkpoint from the base model (before the SFT stage) on a combination of the SFT dataset and all the correct proofs generated during the self-play training whose corresponding statement/conjecture has an empirical pass rate no larger than 1/4. For every statement/conjecture, we randomly keep at most 16 distinct proofs to speedup the training.
\section{Experiments}\label{sec:experiments}
This section presents our implementation details of \algname, the results of Isabelle and Lean experiments, and the ablation studies, followed by examples of generated conjectures.

\subsection{Implementation details}\label{sec:implementation}
\paragraph{Training datasets.} Our primary source of statements without proofs is the de-duplicated LeanWorkbook \citep{ying2024lean}, which contains around 89K Lean4 statements (see Appendix~\ref{app:preprocessing-leanworkbook} for details). For the Isabelle experiments, we translate the Lean4 statements to Isabelle using the DeepSeek V2.5 with few-shot prompting. For the Lean experiments, we combine LeanWorkbook, miniF2F-valid, and ProofNet-valid as the training dataset for STP.

The SFT dataset for the Isabelle experiments is extracted from AFP\footnote{\url{https://www.isa-afp.org/}} and Isabelle built-in files such as HOL. For the Lean experiments, we first sample 32 proofs per statement in LeanWorkbook since our base model, DeepSeek-Prover-V1.5-SFT, is already trained on it, and combine the correct proofs with examples extracted from the proof library Mathlib4 \citep{mathlib} as the SFT dataset.

\paragraph{Periodic refreshing.} With a limited replay buffer, the model may forget some proof skills learned in the SFT stage after many iterations. Therefore, during our \algname training, we periodically re-train the model from the base model on all previously generated correct proofs, following a procedure similar to the final re-training in Section~\ref{sec:checkpointing}. After refreshing, we reset the replay buffer and restart the self-play training using the re-trained model checkpoint.

\paragraph{Verifiers' setup.} To study the scalability of \algname with limited compute, in the Isabelle experiments, we disable the advanced proof tactics \lstinline|sledgehammer, mason, smt, metis, sos|, which require huge CPU compute, to allow more training iterations, sacrificing verification strength and overall performance.
We use PISA \citep{jiang2021lisa} to interact with Isabelle, and enforce a 10s timeout for any proof step and 360s timeout for entire proofs. 
For Lean, we follow \citet{xin2024deepseek}, which allows all proof tactics, and set a 200s timeout and a 15GB memory limit for each proof.

\paragraph{Hyperparameters.} For inference, we cap  the number of generated tokens to 1024, and set the sampling temperature to 0.7 for Llemma-7b and 1.0 for DeepSeek-Prover, following  \citet{dong2024formal, xin2024deepseek}, respectively. For training, we use batch size 2048 and Adam \citep{kingma2014adam} with a constant learning rate of 5e-5 in \algname, and 1e-4 in SFT and final re-training. The discount factors are $\gamma=\exp(-0.001)$ and $\beta=\exp(-0.01)$

In each iteration of \algname, we sample $K=32$ proofs per conjecture/statement. For the expert iteration and parallel sampling, we use $K=64$. Since we maintain the number of generated conjectures per iteration to be at most the number of unproved statements in the given dataset, \algname has the same sample budget as the baseline methods per iteration.

\subsection{Results with Lean}\label{sec:results-lean}

For the Lean experiments, we choose DeepSeek-Prover-V1.5-SFT as our base model, which is trained on proofs collected by expert iteration on a combination of public, such as LeanWorkbook, miniF2F-valid \citep{zheng2021minif2f}, and ProofNet-valid \citep{azerbayev2023proofnet}, and proprietary datasets. We run 48 iterations of \algname and generated 3.6M conjectures, 241M proofs, and 51.3B tokens in total. We use the cumulative pass rate, defined by the fraction of statements proved during the entire training, as the main metric for training progress.
Fig.~\ref{fig:lean-main} plots the cumulative pass rate of \algname and two major baselines, expert iteration, and parallel sampling, on the training dataset LeanWorkbook \citep{ying2024lean}. Expert iteration alternates between generating proofs to the statements in the given dataset and finetuning the model on correct proofs.~(See discussions and comparison about variants of expert iteration in Appendix~\ref{app:expit-details}.) Parallel sampling simply generates proofs with the given model. Fig.~\ref{fig:lean-main} shows that \algname achieves significantly better scaling than expert iteration, which simulates the performance of DeepSeek's model as if it were trained for more iterations.

Since the formal statements in our training dataset, LeanWorkbook, are translated from natural language statements, they are not always provable. In Appendix~\ref{app:leanworkbook-examples}, we randomly select 20 unproved statements from LeanWorkbook and manually assess whether (a) the formal statement is an accurate translation of the natural language statement, and (b) the formal statement itself is correct and provable. We find that 16 out of the 20 statements are translated correctly, but only 7 statements are provable and the remaining 13 statements are unprovable (e.g., due to missing assumptions in the corresponding natural language statement), suggesting that the best possible pass rate on LeanWorkbook, with a 95\% confidence interval, is between 38.7\% and 68.5\%.

In Table~\ref{table:miniF2F-proofnet}, we compare the final re-trained model of STP with prior works on two common benchmarks, miniF2F-test and ProofNet-test, which contain formal statements of high-school level and college level math questions, respectively. Among the whole-proof generation methods, \algname significantly outperforms DeepSeek-Prover-V1.5-RL (which is continuously trained with RL on top of their SFT model) and achieves SoTA performance across various inference-time sample budgets. We also report the performance of the model trained only on LeanWorkbook for 24 iterations, excluding miniF2F-valid and proofnet-valid, demonstrating that the model trained with STP also generalizes to out-of-domain theorems.\footnote{Our base model, DeepSeek-Prover-V1.5-SFT, is trained on miniF2F-valid and ProofNet-valid, though we only run STP on LeanWorkbook in this experiment. The penalization on verification time is also not included in this experiment.}

\begin{table}[tp]
	\centering
	\captionof{table}{Pass rate on miniF2F \citep{zheng2021minif2f} and ProofNet \citep{azerbayev2023proofnet} with different inference-time sample budgets. 
		Our method, \algname, achieves state-of-the-art performance among whole-proof generation methods across various sample budgets.
		For reference, we also include tree search methods, even though they are orthogonal to our main contribution.
		The sample budgets of tree search methods are not fully comparable to that of the whole proof generation because they also use the LLM to process the verifier's internal proof state.
	}
	\label{table:miniF2F-proofnet}
	\begin{small}
		\vspace{-0.05in}
		\begin{tabular}{lcccc}
			\toprule
			\multirow{2}{*}{Method} & Sample budget & Sample budget & \multirow{2}{*}{MiniF2F-test} & \multirow{2}{*}{ProofNet-test} \\ 
			& (\#Proofs) & (\#Steps) & \\
			\toprule 
			\emph{Whole-Proof Generation Methods} \\\midrule
			TheoremLlama \citep{wang2024theoremllama} & 128 & - & 33.6\% & - \\
			DSP \citep{jiang2022draft} & 100 & - & 39.3\% & - \\
			DeepSeek-Prover-V1.5-SFT  & 128 & - & 50.4\% $\pm$ 0.4\% & 15.9\% $\pm$ 0.6\%\\
			~~~\citep{xin2024deepseek} & 3200 & - & 53.3\% $\pm$ 0.5\% & 21.0\% $\pm$ 0.9\% \\
			DeepSeek-Prover-V1.5-RL & 128 & - & 51.6\% $\pm$ 0.5\% & 18.2\% $\pm$ 0.5\%\\
			~~~\citep{xin2024deepseek}& 3200 & - & 54.9\% $\pm$ 0.7\% & 22.0\% $\pm$ 0.5\% \\
			& 25,600 & - & 58.4\% $\pm$ 0.6\% & 23.7\% \\ 
			& {102,400} & - & {60.2\%} & - \\ \midrule
			\text{\algname} & 128 & $1.1$K & 57.2\% $\pm$ 0.6\% & 18.0\% $\pm$ 0.7\% \\
			~~~~\textit{(w/o miniF2F-valid, ProofNet-valid)}& 3200 & $28$K & 61.1\% & 23.1\% \\ %
			\midrule
			\text{\algname} & 128 & 1.3K & \textbf{61.2\% $\pm$ 0.6\%} & \textbf{19.5\% $\pm$ 0.7\%} \\
			& 3200 & 32K & \textbf{65.0\% $\pm$ 0.5\%} & \textbf{23.9\% $\pm$ 0.6\%} \\ 
			& 25,600 & 254K & \textbf{67.6\%} & \textbf{26.9\%} \\
			\bottomrule\toprule
			\emph{Tree Search Methods\footnotemark} \\\midrule
			ReProver \citep{yang2024leandojo} & - & - & 26.5\% & - \\
			PACT \citep{zheng2021minif2f} & - & $8 \times 16 \times 512=66$K &29.2\% & - \\
			GPT-f \citep{polu2022formal} & - & $64 \times 8 \times 512=262$K & 36.6\% & - \\ 
			HTPS \citep{lample2022hypertree} & - & $64 \times 5000=320$K & 41.0\% & - \\ 
			Lean-STaR \citep{lin2024lean} & - & $64 \times 1 \times 50=3.2$K & 46.3\% & - \\
			DeepSeek-Prover-V1.5-RL + RMaxTS\footnotemark & 3200 & - & 55.0\% $\pm$ 0.7\% & 21.5\% $\pm$ 0.8\% \\ 
			~~~\citep{xin2024deepseek} & 25,600 & - & 59.6\% $\pm$ 0.6\% & 25.3\%\\ 
			& {204,800} & - & {63.5\%} & -\\ 
			InternLM2.5-StepProver & - & $4 \times 32 \times 600=77$K & 58.5\% $\pm$ 0.9\% & -\\
			~~~\citep{wu2024internlm2} & - & $16 \times 32 \times 600=307$K & 62.5\% $\pm$ 0.5\% & -\\
			& - & {$256 \times 32 \times 600=4.9$M} & {65.9\%} & -\\
			\bottomrule
		\end{tabular}
	\end{small}
\end{table}
\addtocounter{footnote}{-2}
\stepcounter{footnote}\footnotetext{The \#Steps for tree search methods is typically calculated by \#Independent runs $\times$ \#Tactics generated per search step $\times$ \#Search steps, or \#Independent runs $\times$ \#Search steps. }
\stepcounter{footnote}\footnotetext{DeepSeek-Prover-V1.5-RL + RMaxTS is a tree search method that uses the LLMs to generate complete proofs during the search instead of single proof steps. Therefore, we treat their sample budget as the number of generated proofs instead of steps.}

Table~\ref{table:miniF2F-proofnet} also compares \algname with tree search methods such as InternLM2.5-StepProver \citep{wu2024internlm2}, which use LLMs to generate single proof steps conditioned on the current verifier's proof state and then find a complete proof by best first search or MCTS. The sample budget of these methods are not directly comparable with whole-proof generation methods because (a) the number of steps in a generated proof varies significantly, (b) LLMs in tree search methods need to process additional tokens related to the verifier's proof state, and (c) methods like InternLM2.5-StepProver \citep{wu2024internlm2} require an additional LLM as the value function. Moreover, it's conceivable that tree search methods can also be used with STP, so essentially these are orthogonal methods.
Nonetheless, we compute the total number of proof steps per statement generated by \algname as an proxy for the total number of LLM output tokens for \algname and tree search methods, ignoring the additional compute required by tree search methods to process the verifier's proof state and query the value function. Results in Table~\ref{table:miniF2F-proofnet} indicate that \algname also outperforms prior tree search methods with similar (estimated) inference-time budgets.

As shown in Table~\ref{table:putnam}, on PutnamBench \citep{tsoukalas2024putnambench} which consists of undergraduate-level mathematics competition questions, \algname solves 7 out of 644 problems with 128 samples per problem, and 8 problems with 3200 samples per problem, outperforming the best result of 6 problems in prior works achieved by \citet{wu2024internlm2}.

\begin{figure}[tb]
	\centering
	\includegraphics[width=.33\linewidth]{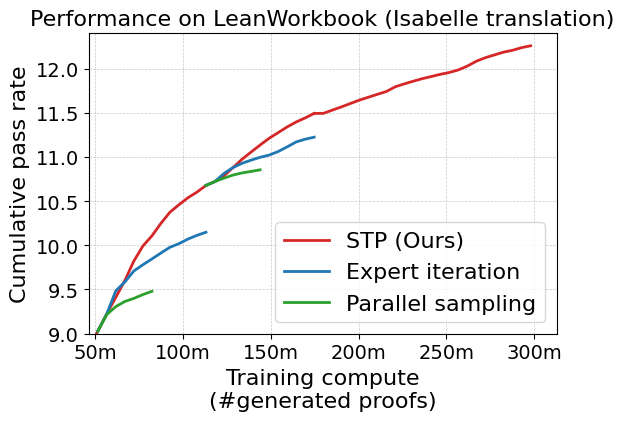}
	\includegraphics[width=.31\linewidth]{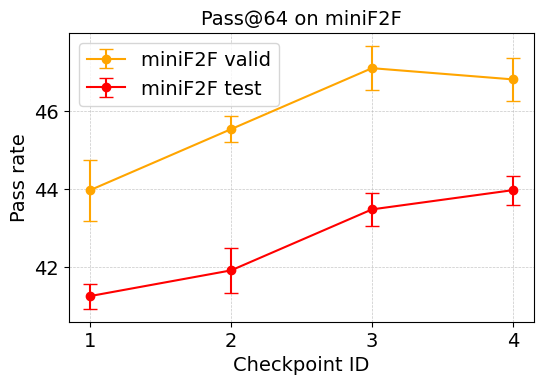}~~~~
	\includegraphics[width=.315\linewidth]{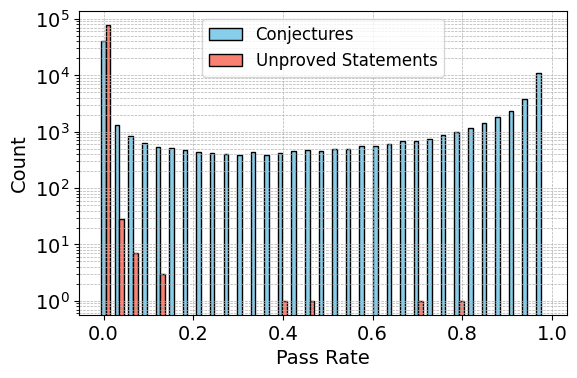}
	\caption{\textbf{Left:} Cumulative pass rate on LeanWorkbook (translated into Isabelle) of \algname, expert iteration, and parallel sampling, started from two checkpoints in \algname training. \algname achieves better scaling starting from both checkpoints. For better visualization, the x-axis starts with 50m in this figure, and we defer the full plot to Fig.~\ref{fig:expit-full} (Right) in Appendix~\ref{app:isabelle-results}. \textbf{Middle:} The performance of our model on miniF2F gradually improves during the training process. Note that our model is not trained on miniF2F valid and we disallow advanced tactics such as \lstinline|sos|. The checkpoints are taken roughly per 68M generated proofs. \textbf{Right:} Histogram of empirical pass rates of generated conjectures and unproved statements in the training dataset at a checkpoint where the cumulative pass rate on LeanWorkbook (Isabelle translation) is 11.4\%. The generated conjectures are significantly more likely to be proved (i.e., has a positive pass rate) than the unproved statements in the dataset, and therefore provide denser training signal. Note that the y-axis is in log scale.}
	\label{fig:isabelle-main}
\end{figure}

\subsection{Results with Isabelle}\label{sec:results-isabelle}

For Isabelle experiments, we start with the Llemma-7b~\citep{azerbayev2023llemma}, math-focused model, and run 58 iterations of \algname to study its scalability. We take several checkpoints during \algname training and then switch to the expert iteration and parallel sampling baselines to study the scalability of the algorithm from checkpoints with various capability. Fig.~\ref{fig:isabelle-main} (Left) compares their cumulative pass rates on LeanWorkbook (Isabelle translation), showing that \algname consistently achieves a better scaling across the training process. The model also gradually improves on  miniF2F over the training process, as shown in Fig.~\ref{fig:isabelle-main} (Middle).

\subsection{Ablation study}\label{sec:ablation}

\paragraph{Generated conjectures provide denser training signals.}
Fig.~\ref{fig:isabelle-main} (Right) shows the histogram of empirical pass rates of the generated conjectures and the unproved statements in LeanWorkbook using a checkpoint in the Isabelle experiment. Only 131 out of 2.5M generated proofs for the 79K unproved statements are correct. As a result, finetuning the model on correct proofs has almost no effect, and thus expert iteration plateaus. In contrast,  generated conjectures by STP offer has higher pass rates and thus more training signals, leading to better scaling.

\paragraph{Re-training with generated conjectures still helps downstream performance.} 
One may hypothesis that the self-play algorithm and generated conjectures only help improve the pass rate on LeanWorkbook. It turns out that in the final re-training stage, it is still beneficial to re-train with the generated conjectures in addition to the successfully proved statements in LeanWorkbook even for performance on miniF2F-test and ProofNet-test---it leads to about 2-3\% performance gain (for pass@128) than re-training only on the latter (See Appendx~\ref{app:lean-results}).

\subsection{Examples of generated conjectures}\label{sec:conjecturing-examples}
In this section, we list three manually selected examples of the generated conjectures at the last iteration of the Lean experiment to demonstrate the quality of generated conjectures.

\paragraph{Example 1.} The generated conjecture says $(1 + x)^{2n} \ge 1 + x^n$ when $n\ge 1$ is an integer and $x\in [0,1]$:
\begin{lstlisting}[frame=single]
theorem lean_workbook_36081' (x : ℝ) (hx : 0 ≤ x ∧ x ≤ 1) : ∀ n :ℕ, n ≥ 1 → (1 + x)^(2*n) ≥ 1 + x^n
\end{lstlisting} The seed statement says $1 + x^2 \le (1 + x)^2$ when $x\in[0,1]$:
\begin{lstlisting}[frame=single]
theorem lean_workbook_36081 (x : ℝ) (hx : 0 ≤ x ∧ x ≤ 1) : 1 + x^2 ≤ (1 + x)^2
\end{lstlisting}
In this case, the conjecture is harder than the original statement but is proved with similar techniques --- expanding the powers of a binomial and then using the fact that $x\ge 0$.

\paragraph{Example 2.} The generated conjecture says $ (x^n - 1) \text{ mod } (x - 1) \le 1$ if $x,n$ are integers:
\begin{lstlisting}[frame=single]
theorem lean_workbook_54038' (x : ℕ) (n : ℕ) (hn : 1 < n) : (x^n - 1) % (x - 1) ≤ 1
\end{lstlisting} 

The seed statement says $n - 1$ divides $n^k - 1$:
\begin{lstlisting}[frame=single]
theorem lean_workbook_54038 (n : ℕ) (k : ℕ) (hn : 2 ≤ n) : n - 1 | n^k - 1
\end{lstlisting}
In this case, our model generates a variant of the original statement by realizing that $b$ mod $a$ equals zero if $a$ divides $b$. This conjecture may help the model connect its proof technique in algebra and number theory. However, the conjecture itself is somewhat unusual and the inequality is not tight. Therefore it is unlikely to be included in any datasets.

\paragraph{Example 3.} The generated conjecture says $\sum_{i\ge 0}((1/4)^i\cdot a)=\frac{a}{1 - 1/4}$ if $0< a\le 1.$
\begin{lstlisting}[frame=single]
theorem lean_workbook_plus_46203' (a : ℝ) (ha : 0 < a ∧ a ≤ 1) :  ∑' (i : ℕ), 1 / 4 ^ i * a = a / (1 - 1 / 4)
\end{lstlisting} The seed statement is a special case where $a=\sqrt{5} / 3$:
\begin{lstlisting}[frame=single]
theorem lean_workbook_plus_46203 :
	∑' k : ℕ, (1 / 4)^k * (Real.sqrt 5 / 4) = (Real.sqrt 5 / 3) 
\end{lstlisting}
In this case, the conjecture generalizes the given statement by replacing \lstinline|Real.sqrt 5 / 4| with a variable \lstinline{a}.

\section{Conclusion}
This paper designs Self-play Theorem Prover (\algname) that simultaneously has two roles, conjecturer and prover. By providing training signals to each other, \algname goes beyond the statements in the given dataset and its performance continuously improves. Our final model significantly outperforms Deepseek-Prover-V1.5 series and achieves state-of-the-art performance among whole-proof generation methods on common formal proof benchmarks.

\subsection*{Acknowledgment}
The authors would like to thank Yinuo Ren, Zhizhou Ren, Woosuk Kwon, David Hall, Huajian Xin and Kaiyue Wen for their helpful discussions. 
The authors would also like to thank the support from NSF RI 2211780, and NSF CIF 2212263, and the Google TPU Research Cloud for the computing resources that enabled these experiments.

\bibliography{new, all}

\begin{thebibliography}{62}
\providecommand{\natexlab}[1]{#1}
\providecommand{\url}[1]{\texttt{#1}}
\expandafter\ifx\csname urlstyle\endcsname\relax
  \providecommand{\doi}[1]{doi: #1}\else
  \providecommand{\doi}{doi: \begingroup \urlstyle{rm}\Url}\fi

\bibitem[AlphaProof(2024)]{alphaproof2024}
AlphaProof.
\newblock Ai achieves silver-medal standard solving international mathematical
  olympiad problems.
\newblock 2024.
\newblock URL
  \url{https://deepmind.google/discover/blog/ai-solves-imo-problems-at-silver-medal-level/}.

\bibitem[Andrychowicz et~al.(2017)Andrychowicz, Wolski, Ray, Schneider, Fong,
  Welinder, McGrew, Tobin, Pieter~Abbeel, and
  Zaremba]{andrychowicz2017hindsight}
Marcin Andrychowicz, Filip Wolski, Alex Ray, Jonas Schneider, Rachel Fong,
  Peter Welinder, Bob McGrew, Josh Tobin, OpenAI Pieter~Abbeel, and Wojciech
  Zaremba.
\newblock Hindsight experience replay.
\newblock \emph{Advances in neural information processing systems}, 30, 2017.

\bibitem[Anthony et~al.(2017)Anthony, Tian, and Barber]{anthony2017thinking}
Thomas Anthony, Zheng Tian, and David Barber.
\newblock Thinking fast and slow with deep learning and tree search.
\newblock \emph{Advances in neural information processing systems}, 30, 2017.

\bibitem[Ayg{\"u}n et~al.(2022)Ayg{\"u}n, Anand, Orseau, Glorot, Mcaleer,
  Firoiu, Zhang, Precup, and Mourad]{aygun2022proving}
Eser Ayg{\"u}n, Ankit Anand, Laurent Orseau, Xavier Glorot, Stephen~M Mcaleer,
  Vlad Firoiu, Lei~M Zhang, Doina Precup, and Shibl Mourad.
\newblock Proving theorems using incremental learning and hindsight experience
  replay.
\newblock In \emph{International Conference on Machine Learning}, pages
  1198--1210. PMLR, 2022.

\bibitem[Azerbayev et~al.(2023{\natexlab{a}})Azerbayev, Piotrowski, Schoelkopf,
  Ayers, Radev, and Avigad]{azerbayev2023proofnet}
Zhangir Azerbayev, Bartosz Piotrowski, Hailey Schoelkopf, Edward~W Ayers,
  Dragomir Radev, and Jeremy Avigad.
\newblock Proofnet: Autoformalizing and formally proving undergraduate-level
  mathematics.
\newblock \emph{arXiv preprint arXiv:2302.12433}, 2023{\natexlab{a}}.

\bibitem[Azerbayev et~al.(2023{\natexlab{b}})Azerbayev, Schoelkopf, Paster,
  Dos~Santos, McAleer, Jiang, Deng, Biderman, and Welleck]{azerbayev2023llemma}
Zhangir Azerbayev, Hailey Schoelkopf, Keiran Paster, Marco Dos~Santos, Stephen
  McAleer, Albert Jiang, Jia Deng, Stella Biderman, and Sean Welleck.
\newblock Llemma: An open language model for mathematics.
\newblock In \emph{The 3rd Workshop on Mathematical Reasoning and AI at
  NeurIPS'23}, 2023{\natexlab{b}}.

\bibitem[Bibel(2013)]{bibel2013automated}
Wolfgang Bibel.
\newblock \emph{Automated theorem proving}.
\newblock Springer Science \& Business Media, 2013.

\bibitem[Colas et~al.(2022)Colas, Karch, Sigaud, and
  Oudeyer]{colas2022autotelic}
C{\'e}dric Colas, Tristan Karch, Olivier Sigaud, and Pierre-Yves Oudeyer.
\newblock Autotelic agents with intrinsically motivated goal-conditioned
  reinforcement learning: a short survey.
\newblock \emph{Journal of Artificial Intelligence Research}, 74:\penalty0
  1159--1199, 2022.

\bibitem[Dong et~al.(2024)Dong, Mahankali, and Ma]{dong2024formal}
Kefan Dong, Arvind Mahankali, and Tengyu Ma.
\newblock Formal theorem proving by rewarding llms to decompose proofs
  hierarchically.
\newblock \emph{arXiv preprint arXiv:2411.01829}, 2024.

\bibitem[Einarsd{\'o}ttir et~al.(2024)Einarsd{\'o}ttir, Alhessi, First, and
  Johansson]{einarsdottir2024lemma}
S{\'o}lr{\'u}n~Halla Einarsd{\'o}ttir, Yousef Alhessi, Emily First, and Moa
  Johansson.
\newblock On lemma conjecturing using neural, symbolic and neuro-symbolic
  approaches.
\newblock 2024.

\bibitem[Guo et~al.(2025)Guo, Yang, Zhang, Song, Zhang, Xu, Zhu, Ma, Wang, Bi,
  et~al.]{guo2025deepseek}
Daya Guo, Dejian Yang, Haowei Zhang, Junxiao Song, Ruoyu Zhang, Runxin Xu,
  Qihao Zhu, Shirong Ma, Peiyi Wang, Xiao Bi, et~al.
\newblock Deepseek-r1: Incentivizing reasoning capability in llms via
  reinforcement learning.
\newblock \emph{arXiv preprint arXiv:2501.12948}, 2025.

\bibitem[Haluptzok et~al.(2022)Haluptzok, Bowers, and
  Kalai]{haluptzok2022language}
Patrick Haluptzok, Matthew Bowers, and Adam~Tauman Kalai.
\newblock Language models can teach themselves to program better.
\newblock \emph{arXiv preprint arXiv:2207.14502}, 2022.

\bibitem[Jaech et~al.(2024)Jaech, Kalai, Lerer, Richardson, El-Kishky, Low,
  Helyar, Madry, Beutel, Carney, et~al.]{jaech2024openai}
Aaron Jaech, Adam Kalai, Adam Lerer, Adam Richardson, Ahmed El-Kishky, Aiden
  Low, Alec Helyar, Aleksander Madry, Alex Beutel, Alex Carney, et~al.
\newblock Openai o1 system card.
\newblock \emph{arXiv preprint arXiv:2412.16720}, 2024.

\bibitem[Jiang et~al.(2022{\natexlab{a}})Jiang, Welleck, Zhou, Li, Liu, Jamnik,
  Lacroix, Wu, and Lample]{jiang2022draft}
Albert~Q Jiang, Sean Welleck, Jin~Peng Zhou, Wenda Li, Jiacheng Liu, Mateja
  Jamnik, Timoth{\'e}e Lacroix, Yuhuai Wu, and Guillaume Lample.
\newblock Draft, sketch, and prove: Guiding formal theorem provers with
  informal proofs.
\newblock \emph{arXiv preprint arXiv:2210.12283}, 2022{\natexlab{a}}.

\bibitem[Jiang et~al.(2023)Jiang, Li, and Jamnik]{jiang2023multilingual}
Albert~Q Jiang, Wenda Li, and Mateja Jamnik.
\newblock Multilingual mathematical autoformalization.
\newblock \emph{arXiv preprint arXiv:2311.03755}, 2023.

\bibitem[Jiang et~al.(2021)Jiang, Li, Han, and Wu]{jiang2021lisa}
Albert~Qiaochu Jiang, Wenda Li, Jesse~Michael Han, and Yuhuai Wu.
\newblock Lisa: Language models of isabelle proofs.
\newblock In \emph{6th Conference on Artificial Intelligence and Theorem
  Proving}, pages 378--392, 2021.

\bibitem[Jiang et~al.(2022{\natexlab{b}})Jiang, Li, Tworkowski, Czechowski,
  Odrzyg{\'o}{\'z}d{\'z}, Mi{\l}o{\'s}, Wu, and Jamnik]{jiang2022thor}
Albert~Qiaochu Jiang, Wenda Li, Szymon Tworkowski, Konrad Czechowski, Tomasz
  Odrzyg{\'o}{\'z}d{\'z}, Piotr Mi{\l}o{\'s}, Yuhuai Wu, and Mateja Jamnik.
\newblock Thor: Wielding hammers to integrate language models and automated
  theorem provers.
\newblock \emph{Advances in Neural Information Processing Systems},
  35:\penalty0 8360--8373, 2022{\natexlab{b}}.

\bibitem[Johansson and Smallbone(2023)]{johansson2023exploring}
Moa Johansson and Nicholas Smallbone.
\newblock Exploring mathematical conjecturing with large language models.
\newblock 2023.

\bibitem[Kaliszyk et~al.(2018)Kaliszyk, Urban, Michalewski, and
  Ol{\v{s}}{\'a}k]{kaliszyk2018reinforcement}
Cezary Kaliszyk, Josef Urban, Henryk Michalewski, and Miroslav Ol{\v{s}}{\'a}k.
\newblock Reinforcement learning of theorem proving.
\newblock \emph{Advances in Neural Information Processing Systems}, 31, 2018.

\bibitem[Kingma and Ba(2014)]{kingma2014adam}
Diederik~P Kingma and Jimmy Ba.
\newblock Adam: A method for stochastic optimization.
\newblock \emph{arXiv preprint arXiv:1412.6980}, 2014.

\bibitem[Kwon et~al.(2023)Kwon, Li, Zhuang, Sheng, Zheng, Yu, Gonzalez, Zhang,
  and Stoica]{kwon2023efficient}
Woosuk Kwon, Zhuohan Li, Siyuan Zhuang, Ying Sheng, Lianmin Zheng, Cody~Hao Yu,
  Joseph Gonzalez, Hao Zhang, and Ion Stoica.
\newblock Efficient memory management for large language model serving with
  pagedattention.
\newblock In \emph{Proceedings of the 29th Symposium on Operating Systems
  Principles}, pages 611--626, 2023.

\bibitem[Lample et~al.(2022)Lample, Lacroix, Lachaux, Rodriguez, Hayat, Lavril,
  Ebner, and Martinet]{lample2022hypertree}
Guillaume Lample, Timothee Lacroix, Marie-Anne Lachaux, Aurelien Rodriguez,
  Amaury Hayat, Thibaut Lavril, Gabriel Ebner, and Xavier Martinet.
\newblock Hypertree proof search for neural theorem proving.
\newblock \emph{Advances in neural information processing systems},
  35:\penalty0 26337--26349, 2022.

\bibitem[Li et~al.(2023)Li, Allal, Zi, Muennighoff, Kocetkov, Mou, Marone,
  Akiki, Li, Chim, et~al.]{li2023starcoder}
Raymond Li, Loubna~Ben Allal, Yangtian Zi, Niklas Muennighoff, Denis Kocetkov,
  Chenghao Mou, Marc Marone, Christopher Akiki, Jia Li, Jenny Chim, et~al.
\newblock Starcoder: may the source be with you!
\newblock \emph{arXiv preprint arXiv:2305.06161}, 2023.

\bibitem[Lin et~al.(2024)Lin, Sun, Yang, and Welleck]{lin2024lean}
Haohan Lin, Zhiqing Sun, Yiming Yang, and Sean Welleck.
\newblock Lean-star: Learning to interleave thinking and proving.
\newblock \emph{arXiv preprint arXiv:2407.10040}, 2024.

\bibitem[Loveland(2016)]{loveland2016automated}
Donald~W Loveland.
\newblock \emph{Automated theorem proving: A logical basis}.
\newblock Elsevier, 2016.

\bibitem[Lu et~al.(2024)Lu, Wan, Liu, Huang, Xiong, Liu, Shen, Jin, Zhang,
  Wang, et~al.]{lu2024process}
Jianqiao Lu, Yingjia Wan, Zhengying Liu, Yinya Huang, Jing Xiong, Chengwu Liu,
  Jianhao Shen, Hui Jin, Jipeng Zhang, Haiming Wang, et~al.
\newblock Process-driven autoformalization in lean 4.
\newblock \emph{arXiv preprint arXiv:2406.01940}, 2024.

\bibitem[mathlib Community(2020)]{mathlib}
The mathlib Community.
\newblock The lean mathematical library.
\newblock In \emph{Proceedings of the 9th ACM SIGPLAN International Conference
  on Certified Programs and Proofs}, CPP 2020, page 367–381, New York, NY,
  USA, 2020. Association for Computing Machinery.
\newblock ISBN 9781450370974.
\newblock \doi{10.1145/3372885.3373824}.
\newblock URL \url{https://doi.org/10.1145/3372885.3373824}.

\bibitem[Moura and Ullrich(2021)]{moura2021lean}
Leonardo~de Moura and Sebastian Ullrich.
\newblock The lean 4 theorem prover and programming language.
\newblock In \emph{Automated Deduction--CADE 28: 28th International Conference
  on Automated Deduction, Virtual Event, July 12--15, 2021, Proceedings 28},
  pages 625--635. Springer, 2021.

\bibitem[Nijkamp et~al.(2022)Nijkamp, Pang, Hayashi, Tu, Wang, Zhou, Savarese,
  and Xiong]{nijkamp2022codegen}
Erik Nijkamp, Bo~Pang, Hiroaki Hayashi, Lifu Tu, Huan Wang, Yingbo Zhou, Silvio
  Savarese, and Caiming Xiong.
\newblock Codegen: An open large language model for code with multi-turn
  program synthesis.
\newblock \emph{arXiv preprint arXiv:2203.13474}, 2022.

\bibitem[Nipkow et~al.(2002)Nipkow, Wenzel, and Paulson]{nipkow2002isabelle}
Tobias Nipkow, Markus Wenzel, and Lawrence~C Paulson.
\newblock \emph{Isabelle/HOL: a proof assistant for higher-order logic}.
\newblock Springer, 2002.

\bibitem[Parker-Holder et~al.(2022)Parker-Holder, Jiang, Dennis, Samvelyan,
  Foerster, Grefenstette, and Rockt{\"a}schel]{parker2022evolving}
Jack Parker-Holder, Minqi Jiang, Michael Dennis, Mikayel Samvelyan, Jakob
  Foerster, Edward Grefenstette, and Tim Rockt{\"a}schel.
\newblock Evolving curricula with regret-based environment design.
\newblock In \emph{International Conference on Machine Learning}, pages
  17473--17498. PMLR, 2022.

\bibitem[Plaat et~al.(2024)Plaat, Wong, Verberne, Broekens, van Stein, and
  Back]{plaat2024reasoning}
Aske Plaat, Annie Wong, Suzan Verberne, Joost Broekens, Niki van Stein, and
  Thomas Back.
\newblock Reasoning with large language models, a survey.
\newblock \emph{arXiv preprint arXiv:2407.11511}, 2024.

\bibitem[Poesia et~al.(2024)Poesia, Broman, Haber, and
  Goodman]{poesia2024learning}
Gabriel Poesia, David Broman, Nick Haber, and Noah~D Goodman.
\newblock Learning formal mathematics from intrinsic motivation.
\newblock \emph{arXiv preprint arXiv:2407.00695}, 2024.

\bibitem[Polu et~al.(2022)Polu, Han, Zheng, Baksys, Babuschkin, and
  Sutskever]{polu2022formal}
Stanislas Polu, Jesse~Michael Han, Kunhao Zheng, Mantas Baksys, Igor
  Babuschkin, and Ilya Sutskever.
\newblock Formal mathematics statement curriculum learning.
\newblock \emph{arXiv preprint arXiv:2202.01344}, 2022.

\bibitem[Portelas et~al.(2020)Portelas, Colas, Weng, Hofmann, and
  Oudeyer]{portelas2020automatic}
R{\'e}my Portelas, C{\'e}dric Colas, Lilian Weng, Katja Hofmann, and
  Pierre-Yves Oudeyer.
\newblock Automatic curriculum learning for deep rl: A short survey.
\newblock \emph{arXiv preprint arXiv:2003.04664}, 2020.

\bibitem[Pourcel et~al.(2024{\natexlab{a}})Pourcel, Carta, Kova{\v{c}}, and
  Oudeyer]{pourcel2024autotelic}
Guillaume Pourcel, Thomas Carta, Grgur Kova{\v{c}}, and Pierre-Yves Oudeyer.
\newblock Autotelic llm-based exploration for goal-conditioned rl.
\newblock In \emph{Intrinsically Motivated Open-ended Learning Workshop at
  NeurIPS 2024}, 2024{\natexlab{a}}.

\bibitem[Pourcel et~al.(2024{\natexlab{b}})Pourcel, Colas, Molinaro, Oudeyer,
  and Teodorescu]{pourcel2024aces}
Julien Pourcel, C{\'e}dric Colas, Gaia Molinaro, Pierre-Yves Oudeyer, and
  Laetitia Teodorescu.
\newblock Aces: generating diverse programming puzzles with autotelic language
  models and semantic descriptors.
\newblock \emph{Neurips}, 2024{\natexlab{b}}.

\bibitem[Shao et~al.(2024)Shao, Wang, Zhu, Xu, Song, Zhang, Li, Wu, and
  Guo]{shao2024deepseekmath}
Zhihong Shao, Peiyi Wang, Qihao Zhu, Runxin Xu, Junxiao Song, Mingchuan Zhang,
  YK~Li, Y~Wu, and Daya Guo.
\newblock Deepseekmath: Pushing the limits of mathematical reasoning in open
  language models.
\newblock \emph{arXiv preprint arXiv:2402.03300}, 2024.

\bibitem[Shinn et~al.(2023)Shinn, Cassano, Labash, Gopinath, Narasimhan, and
  Yao]{shinn2023reflexion}
Noah Shinn, Federico Cassano, Beck Labash, Ashwin Gopinath, Karthik Narasimhan,
  and Shunyu Yao.
\newblock Reflexion: Language agents with verbal reinforcement learning.(2023).
\newblock \emph{arXiv preprint cs.AI/2303.11366}, 2023.

\bibitem[Silver et~al.(2016)Silver, Huang, Maddison, Guez, Sifre, van~den
  Driessche, Schrittwieser, Antonoglou, Panneershelvam, Lanctot, Dieleman,
  Grewe, Nham, Kalchbrenner, Sutskever, Lillicrap, Leach, Kavukcuoglu, Graepel,
  and Hassabis]{silver2017mastering}
David Silver, Aja Huang, Christopher~J. Maddison, Arthur Guez, Laurent Sifre,
  George van~den Driessche, Julian Schrittwieser, Ioannis Antonoglou, Veda
  Panneershelvam, Marc Lanctot, Sander Dieleman, Dominik Grewe, John Nham, Nal
  Kalchbrenner, Ilya Sutskever, Timothy Lillicrap, Madeleine Leach, Koray
  Kavukcuoglu, Thore Graepel, and Demis Hassabis.
\newblock Mastering the game of {Go} with deep neural networks and tree search.
\newblock \emph{Nature}, 529\penalty0 (7676):\penalty0 484--503, 2016.

\bibitem[Sun et~al.(2018)Sun, Gordon, Boots, and Bagnell]{sun2018dual}
Wen Sun, Geoffrey~J Gordon, Byron Boots, and J~Bagnell.
\newblock Dual policy iteration.
\newblock \emph{Advances in Neural Information Processing Systems}, 31, 2018.

\bibitem[Teodorescu et~al.(2023)Teodorescu, Colas, Bowers, Carta, and
  Oudeyer]{teodorescu2023codeplay}
Laetitia Teodorescu, C{\'e}dric Colas, Matthew Bowers, Thomas Carta, and
  Pierre-Yves Oudeyer.
\newblock Codeplay: Autotelic learning through collaborative self-play in
  programming environments.
\newblock In \emph{IMOL 2023-Intrinsically Motivated Open-ended Learning
  workshop at NeurIPS 2023}, 2023.

\bibitem[Touvron et~al.(2023)Touvron, Martin, Stone, Albert, Almahairi, Babaei,
  Bashlykov, Batra, Bhargava, Bhosale, et~al.]{touvron2023llama}
Hugo Touvron, Louis Martin, Kevin Stone, Peter Albert, Amjad Almahairi, Yasmine
  Babaei, Nikolay Bashlykov, Soumya Batra, Prajjwal Bhargava, Shruti Bhosale,
  et~al.
\newblock Llama 2: Open foundation and fine-tuned chat models.
\newblock \emph{arXiv preprint arXiv:2307.09288}, 2023.

\bibitem[Trinh and Luong(2024)]{trinh2024alphageometry}
Trieu Trinh and Thang Luong.
\newblock Alphageometry: An olympiad-level ai system for geometry.
\newblock \emph{Google DeepMind}, 17, 2024.

\bibitem[Tsoukalas et~al.(2024)Tsoukalas, Lee, Jennings, Xin, Ding, Jennings,
  Thakur, and Chaudhuri]{tsoukalas2024putnambench}
George Tsoukalas, Jasper Lee, John Jennings, Jimmy Xin, Michelle Ding, Michael
  Jennings, Amitayush Thakur, and Swarat Chaudhuri.
\newblock Putnambench: Evaluating neural theorem-provers on the putnam
  mathematical competition.
\newblock \emph{arXiv preprint arXiv:2407.11214}, 2024.

\bibitem[Urban and Jakub{\r{u}}v(2020)]{urban2020first}
Josef Urban and Jan Jakub{\r{u}}v.
\newblock First neural conjecturing datasets and experiments.
\newblock In \emph{Intelligent Computer Mathematics: 13th International
  Conference, CICM 2020, Bertinoro, Italy, July 26--31, 2020, Proceedings 13},
  pages 315--323. Springer, 2020.

\bibitem[Wang et~al.(2023)Wang, Xin, Zheng, Li, Liu, Cao, Huang, Xiong, Shi,
  Xie, et~al.]{wang2023lego}
Haiming Wang, Huajian Xin, Chuanyang Zheng, Lin Li, Zhengying Liu, Qingxing
  Cao, Yinya Huang, Jing Xiong, Han Shi, Enze Xie, et~al.
\newblock Lego-prover: Neural theorem proving with growing libraries.
\newblock \emph{arXiv preprint arXiv:2310.00656}, 2023.

\bibitem[Wang and Deng(2020)]{wang2020learning}
Mingzhe Wang and Jia Deng.
\newblock Learning to prove theorems by learning to generate theorems.
\newblock In \emph{Proceedings of the 34th International Conference on Neural
  Information Processing Systems}, pages 18146--18157, 2020.

\bibitem[Wang et~al.(2024)Wang, Zhang, Jia, Pan, Diao, Pi, and
  Zhang]{wang2024theoremllama}
Ruida Wang, Jipeng Zhang, Yizhen Jia, Rui Pan, Shizhe Diao, Renjie Pi, and Tong
  Zhang.
\newblock Theoremllama: Transforming general-purpose llms into lean4 experts.
\newblock \emph{arXiv preprint arXiv:2407.03203}, 2024.

\bibitem[Wu et~al.(2021)Wu, Norrish, Walder, and Dezfouli]{wu2021tacticzero}
Minchao Wu, Michael Norrish, Christian Walder, and Amir Dezfouli.
\newblock Tacticzero: Learning to prove theorems from scratch with deep
  reinforcement learning.
\newblock \emph{Advances in Neural Information Processing Systems},
  34:\penalty0 9330--9342, 2021.

\bibitem[Wu et~al.(2020)Wu, Jiang, Ba, and Grosse]{wu2020int}
Yuhuai Wu, Albert~Qiaochu Jiang, Jimmy Ba, and Roger Grosse.
\newblock Int: An inequality benchmark for evaluating generalization in theorem
  proving.
\newblock \emph{arXiv preprint arXiv:2007.02924}, 2020.

\bibitem[Wu et~al.(2024)Wu, Huang, Zhou, Ying, Wang, Lin, and
  Chen]{wu2024internlm2}
Zijian Wu, Suozhi Huang, Zhejian Zhou, Huaiyuan Ying, Jiayu Wang, Dahua Lin,
  and Kai Chen.
\newblock Internlm2. 5-stepprover: Advancing automated theorem proving via
  expert iteration on large-scale lean problems.
\newblock \emph{arXiv preprint arXiv:2410.15700}, 2024.

\bibitem[Xin et~al.(2024{\natexlab{a}})Xin, Guo, Shao, Ren, Zhu, Liu, Ruan, Li,
  and Liang]{xin2024deepseeka}
Huajian Xin, Daya Guo, Zhihong Shao, Zhizhou Ren, Qihao Zhu, Bo~Liu, Chong
  Ruan, Wenda Li, and Xiaodan Liang.
\newblock Deepseek-prover: Advancing theorem proving in llms through
  large-scale synthetic data.
\newblock \emph{arXiv preprint arXiv:2405.14333}, 2024{\natexlab{a}}.

\bibitem[Xin et~al.(2024{\natexlab{b}})Xin, Ren, Song, Shao, Zhao, Wang, Liu,
  Zhang, Lu, Du, et~al.]{xin2024deepseek}
Huajian Xin, ZZ~Ren, Junxiao Song, Zhihong Shao, Wanjia Zhao, Haocheng Wang,
  Bo~Liu, Liyue Zhang, Xuan Lu, Qiushi Du, et~al.
\newblock Deepseek-prover-v1.5: Harnessing proof assistant feedback for
  reinforcement learning and monte-carlo tree search.
\newblock \emph{arXiv preprint arXiv:2408.08152}, 2024{\natexlab{b}}.

\bibitem[Yang et~al.(2024{\natexlab{a}})Yang, Poesia, He, Li, Lauter,
  Chaudhuri, and Song]{yang2024formal}
Kaiyu Yang, Gabriel Poesia, Jingxuan He, Wenda Li, Kristin Lauter, Swarat
  Chaudhuri, and Dawn Song.
\newblock Formal mathematical reasoning: A new frontier in ai.
\newblock \emph{arXiv preprint arXiv:2412.16075}, 2024{\natexlab{a}}.

\bibitem[Yang et~al.(2024{\natexlab{b}})Yang, Swope, Gu, Chalamala, Song, Yu,
  Godil, Prenger, and Anandkumar]{yang2024leandojo}
Kaiyu Yang, Aidan Swope, Alex Gu, Rahul Chalamala, Peiyang Song, Shixing Yu,
  Saad Godil, Ryan~J Prenger, and Animashree Anandkumar.
\newblock Leandojo: Theorem proving with retrieval-augmented language models.
\newblock \emph{Advances in Neural Information Processing Systems}, 36,
  2024{\natexlab{b}}.

\bibitem[Yao et~al.(2022)Yao, Zhao, Yu, Du, Shafran, Narasimhan, and
  Cao]{yao2022react}
Shunyu Yao, Jeffrey Zhao, Dian Yu, Nan Du, Izhak Shafran, Karthik Narasimhan,
  and Yuan Cao.
\newblock React: Synergizing reasoning and acting in language models.
\newblock \emph{arXiv preprint arXiv:2210.03629}, 2022.

\bibitem[Ye et~al.(2024)Ye, Agarwal, Liu, Joshi, Velury, Le, Tan, and
  Liu]{ye2024evolving}
Ziyu Ye, Rishabh Agarwal, Tianqi Liu, Rishabh Joshi, Sarmishta Velury, Quoc~V
  Le, Qijun Tan, and Yuan Liu.
\newblock Evolving alignment via asymmetric self-play.
\newblock \emph{arXiv preprint arXiv:2411.00062}, 2024.

\bibitem[Ying et~al.(2024)Ying, Wu, Geng, Wang, Lin, and Chen]{ying2024lean}
Huaiyuan Ying, Zijian Wu, Yihan Geng, Jiayu Wang, Dahua Lin, and Kai Chen.
\newblock Lean workbook: A large-scale lean problem set formalized from natural
  language math problems.
\newblock \emph{arXiv preprint arXiv:2406.03847}, 2024.

\bibitem[Zhang et~al.(2023)Zhang, Lehman, Stanley, and Clune]{zhang2023omni}
Jenny Zhang, Joel Lehman, Kenneth Stanley, and Jeff Clune.
\newblock Omni: Open-endedness via models of human notions of interestingness.
\newblock \emph{arXiv preprint arXiv:2306.01711}, 2023.

\bibitem[Zheng et~al.(2023)Zheng, Wang, Xie, Liu, Sun, Xin, Shen, Li, and
  Li]{zheng2023lyra}
Chuanyang Zheng, Haiming Wang, Enze Xie, Zhengying Liu, Jiankai Sun, Huajian
  Xin, Jianhao Shen, Zhenguo Li, and Yu~Li.
\newblock Lyra: Orchestrating dual correction in automated theorem proving.
\newblock \emph{arXiv preprint arXiv:2309.15806}, 2023.

\bibitem[Zheng et~al.(2021)Zheng, Han, and Polu]{zheng2021minif2f}
Kunhao Zheng, Jesse~Michael Han, and Stanislas Polu.
\newblock minif2f: a cross-system benchmark for formal olympiad-level
  mathematics.
\newblock In \emph{International Conference on Learning Representations}, 2021.

\end{thebibliography}
\bibliographystyle{plainnat}

\newpage
\appendix
\onecolumn

\section{Additional Implementation Details}
In this section, we list the missing implementation details.

\subsection{Examples of inputs and outputs of our model}\label{app:examples}
Here we present some concrete examples to demonstrate the input and output formats of our model.

\paragraph{Examples of the conjecturer.} In the following, we show examples of the conjecturer's inputs and outputs. Note that \lstinline|<lemma>,<easy theorem>,<hard theorem>,</hard theorem>| are the formatting tokens, and the system prompt is the first three lines in the input examples.

\noindent \textit{Input}:
\begin{lstlisting}[frame=single]
Complete the following Lean 4 code:

```lean4
<lemma>
lemma sq_nonneg (a : α) : 0 ≤ a ^ 2
<easy theorem>
theorem lean_workbook_9742 (a b c : ℝ) (ha : a ≥ 0) (hb : b ≥ 0) (hc : c ≥ 0) (hab : a + b + c = 1) : a^2 + b^2 + c^2 ≤ 1 / 4 + a^3 + b^3 + c^3  :=  by
nlinarith [sq_nonneg (a - 1 / 2), sq_nonneg (b - 1 / 2), sq_nonneg (c - 1 / 2),
mul_nonneg ha hb, mul_nonneg hb hc, mul_nonneg ha hc, hab]
<hard theorem>
theorem
\end{lstlisting}
\noindent \textit{Output}:
\begin{lstlisting}[frame=single]
lean_workbook_9229 (a b c : ℝ) (ha : 0 < a ∧ 0 < b ∧ 0 < c) (hab : a + b + c = 1) : a^3 / (a * b) + b^3 / (b * c) + c^3 / (c * a) ≥ 1 - a^2 - b^2 - c^2</hard theorem>
```
\end{lstlisting}

In the following example, the lemma is replaced by a trivial statement, and the model can generate conjectures without focusing on any particular direction (see Section~\ref{sec:implementation}).

\noindent \textit{Input}:
\begin{lstlisting}[frame=single]
Complete the following Lean 4 code:

```lean4
<lemma>
theorem true: True
<easy theorem>
theorem lean_workbook_980 (a b : ℝ) (ha : 0 < a) (hb : 0 < b) (hab : 1/a + 1/b = 1) : 1/(a + 1) + 2/(2 * b + 1) ≤ 3/4  :=  by
field_simp [ha.ne', hb.ne', ha.ne', hb.ne'] at hab
field_simp [ha.ne', hb.ne', ha.ne', hb.ne']
rw [div_le_div_iff (by positivity) (by positivity)]
nlinarith [sq_nonneg (a - b), sq_nonneg (a - 2 * b), sq_nonneg (2 * a - b), hab]
<hard theorem>
theorem
\end{lstlisting}
\noindent \textit{Output}:
\begin{lstlisting}[frame=single]
lean_workbook_982 (a b c : ℝ) (ha : 0 < a) (hb : 0 < b) (hc : 0 < c) (hab : a + b + c = 1) :
a / a ^ 2 + b / b ^ 2 + c / c ^ 2 ≥ 3 / 4 ∧ 1 / (a + b + c) ^ 2 + 1 / (b + c + a) ^ 2 + 1 / (c + a + b) ^ 2 ≥ 3 / 4</hard theorem>
```
\end{lstlisting}

\paragraph{Examples of the prover.} In the following, we show an example of the prover's inputs and outputs.

\noindent \textit{Input}:
\begin{lstlisting}[frame=single]
Complete the following Lean 4 code:

```lean4
theorem lean_workbook_52370 (p q : ℕ) (hp : Nat.Prime p) (hq : Nat.Prime q) (hpq: p >= q) (hq5 : q >= 5) : 24 | (p^2 - q^2)  :=  by
\end{lstlisting}
\noindent \textit{Output}:
\begin{lstlisting}[frame=single]
have : p^2 % 24 = (p % 24)^2 % 24 := by
rw [Nat.pow_mod]
have : q^2 % 24 = (q % 24)^2 % 24 := by
rw [Nat.pow_mod]
rw [← Nat.mod_add_div p 24, ← Nat.mod_add_div q 24]
ring_nf
norm_num
have : p % 24 = 1 ∨ p % 24 = 5 ∨ p % 24 = 7 ∨ p % 24 = 11 ∨ p % 24 = 13 ∨ p % 24 = 17 ∨ p % 24 = 19 ∨ p % 24 = 23 := by
have := hp.eq_one_or_self_of_dvd 2
have := hp.eq_one_or_self_of_dvd 3
have := hp.eq_one_or_self_of_dvd 4
have := hp.eq_one_or_self_of_dvd 5
omega
have : q % 24 = 1 ∨ q % 24 = 5 ∨ q % 24 = 7 ∨ q % 24 = 11 ∨ q % 24 = 13 ∨ q % 24 = 17 ∨ q % 24 = 19 ∨ q % 24 = 23 := by
have := hq.eq_one_or_self_of_dvd 2
have := hq.eq_one_or_self_of_dvd 3
have := hq.eq_one_or_self_of_dvd 4
have := hq.eq_one_or_self_of_dvd 5
omega
rcases this with (h | h | h | h | h | h | h | h) <;> rcases this with (h' | h' | h' | h' | h' | h' | h' | h') <;>
simp [h, h', pow_two, Nat.mul_mod, Nat.add_mod, Nat.mod_mod, Nat.mod_self, zero_add,
Nat.zero_mul, Nat.mod_one, Nat.one_mul, Nat.mod_add_mod]
<;> omega
```
\end{lstlisting}

\subsection{Pseudo-code for selecting the conjecturer's inputs}\label{app:step1-pseudo-code}
In the following, we present the pseudo-code for selecting the conjecturing imports. Recall that the input for the conjecturer consists of a statement, its proof, and a lemma used in the proof (c.f., Section~\ref{sec:sft}). In Step 1, we construct the prompts by taking the correct proofs to statements in the given dataset, and extract the lemmas used in the proof by the formal verifiers. We also allow the model to propose conjectures without focusing on any particular lemma, which is implemented by replacing the lemma statement with a fixed trivial statement in the prompt (see Appendix~\ref{app:examples} for concrete examples). Finally, we de-duplicate the inputs by the (statement, lemma) pair. After generating the conjectures, we randomly select a subset whose size does not exceed the number of remaining unproved statements in the given dataset, so that the prover's sample budget is distributed equally between the conjectures and the statements.

We run two heuristic methods to ensure the diversity of the inputs. First, we make sure that each lemma $l$ appears at most $0.1n$ times in the inputs because we observe that some lemmas (e.g., \lstinline|sq_nonneg, mul_self_nonneg|) are much more likely to be included. Second, we make sure that every statement-lemma pair only appear at most once in the prompt, even if there are multiple correct proofs.

Alg.~\ref{alg:step1-inputs} shows the complete pseudo-code for selecting the conjecturer's inputs.

\begin{algorithm}[htp]
	\caption{Prepare inputs for the conjecturer.}
	\label{alg:step1-inputs}
	\begin{algorithmic}[1]
		\State {\bfseries Input:} a list of statements and proofs $L=\{(t_i, p_i)\}_{i=1}^{n}$.
		\State Initialize prompt list $P=[].$
		\For{$(t,p)\in L$}
		\State Parse the proof and get the set of used lemmas $S$.
		\State With probability 0.5, add the trivial lemma to $S$.
		\State For every lemma $l\in S$, add $(t,p,l)$ to the prompt list $P$.
		\EndFor
		\For{$l\in S$}
		\If{$\sum_{(t',p',l')\in P}\ind{l=l'} > 0.1n$}
		\State Randomly keep at most $0.1n$ prompts with lemma $l$ in $P$.
		\EndIf
		\EndFor
		\State De-duplicate $P$ randomly so that every (statement, lemma) pair $(t,l)$ appears at most once.
		\State {\bfseries Return:} de-duplicated list of prompts $P$.
	\end{algorithmic}
\end{algorithm}

\subsection{Pseudo-code for preparing the conjecturer dataset.}\label{app:step4-pesudo-code}
The pseudo-code for preparing the conjecturer dataset is shown in Alg.~\ref{alg:conjecture-dataset}. The motivations and explanations of each step in Alg.~\ref{alg:conjecture-dataset} can be found in Section~\ref{sec:stp-loop}.

\begin{algorithm}[htp]
	\caption{Prepare the conjecturer dataset.}
	\label{alg:conjecture-dataset}
	\begin{algorithmic}[1]
		\State {\bfseries Input:} a list of (seed statement, proof of the seed statement, lemma, generated conjecture, generated proof of the conjecture) tuples $\calD=\{(t_i, p^t_i, l_i, c_i, p^c_i)\}_{i=1,\cdots,n}$, and unproved statements $Q=\{t_j\}_{j=1,\cdots,m}.$
		\State For each conjecture $c$, compute its empirical pass rate  $$\textstyle{\hat{P}(c)\defeq (\#\{i:c_i=c,p^c_i\text{ is correct}\}) / (\#\{i:c_i=c\}).}$$
		\State Select conjecturing examples that (a) have low but positive pass rates, and (b) the lemma $l$ is used in the proof $p^c$:
		\begin{align*}\overline{\calD}=\{(t, p^t, l, c) \mid &(t, p^t, l, c, p^c)\in \calD, \hat{P}(c)\in (0,1/4],\\&\quad p^c\text{ is correct}, l\text{ is used in }p^c\}.\end{align*}
		\State De-duplicate $\overline{\calD}$ based on the conjecture $c$.
		\State Compute the elegancy score 
		\begin{align*} E(c)\defeq \frac{\min\{ \len(p_i^c): 1\le i\le n, p_i^c\text{ is correct}, c_i=c\}}{\len(c)} \end{align*}
		\State Let $\kappa$ be the 20\%-quantile of $E(c)$ for conjectures in $\overline{\calD}$.
		\State Apply elegancy filter: $\widetilde{\calD}=\{(t, p^t, l, c)\in \overline\calD\mid E(c) \ge \kappa\}.$
		\State Find a distribution $P$ supported on the conjectures in $\widetilde{\calD}$ that minimizes the Wasserstein distance $W(P, Q)$ (Alg.~\ref{alg:matching}).
		\State \textbf{Return:} $\widetilde{\calD}$ re-weighted by the density of $P$.
	\end{algorithmic}
\end{algorithm}

\subsection{Pre-processing LeanWorkbook}\label{app:preprocessing-leanworkbook}
LeanWorkbook is a dataset that contains statements translated from natural language math statements (a.k.a., auto-formalization). The original dataset contains 140K (natural language statement, formal statement) pairs.

We de-duplicate the LeanWorkbook dataset by keeping only one formal statements per natural language statement. After de-duplication, we get 89,221 formal Lean4 statements as our existing dataset w/o proofs for Lean experiments.

For the Isabelle experiments, we translate the Lean4 statements to Isabelle using DeepSeek-V2.5 API with few-shot examples. The prompt to the model is listed below.

\begin{lstlisting}[frame=single,language=]
Please translate the following lean statement into Isabelle. Please make sure that
1. All the variables are well-typed.
2. All the functions are correctly translated into the corresponding Isabelle functions.
3. All the symbols are correctly translated into corresponding Isabelle symbols.
4. Please directly output the translation without explanation.

Here are some hints for the translation:
1. In Isabelle, the second operand of the operator ^ should be integer. For real numbers, please use powr instead.
2. Please define the types of numerals.
3. `Real.logb x y` should be translated to `log x y`.
4. `Real.sqrt x` should be translated to `sqrt x`.
5. Variables with subscripts should be disallowed. For any variable names of form a_b, translate it to ab.
6. Please translate superscripts to the corresponding exponential form. For example, x⁻¹ should be translated to (x powr -1).
7. `a | b` should be translated to `a dvd b`.
8. `x ≡ y [ZMOD p]` should be translated to `x mod p = y mod p`.
9. `x ∈ zmod p` should represent that x is nat and x < p.

## Input:
```lean
theorem lean_workbook_50 (a b c : ℝ) 
(ha : a ≥ 0 ∧ b ≥ 0 ∧ c ≥ 0) 
(hab : a + b + c = 3) 
: a^3 + b^3 + c^3 + 216 * (a * b + b * c + c * a) / (24 + a * b + b * c + c * a) ≤ 27  :=  by sorry
```                                     

## Output:
```Isabelle
theorem lean_workbook_50:
fixes a b c :: real
assumes "a ≥ 0 ∧ b ≥ 0 ∧ c ≥ 0"
assumes "a + b + c = 3"
shows "a^3 + b^3 + c^3 + 216 * (a * b + b * c + c * a) / (24 + a * b + b * c + c * a) ≤ 27"
sorry                                 
```

## Input:
```lean
{}
```

## Output:
\end{lstlisting}

\subsection{Re-weighting the conjecturing dataset}\label{app:wasserstein}
In this section, we describe the motivations and implementation details of the re-weighting method for the conjecturing dataset. 

\paragraph{Motivation.} In our early experiments, we observe that the generated conjectures tend to have mode collapse issue after several iterations of self-play training. For example, the generated conjectures are mostly about algebraic manipulations even when the seed statements contain questions about, for example, number theory. This is partly because the LeanWorkbook dataset contains a significant portion of inequality questions.

Therefore, in addition to the particular conjecturing format where we require that the proof of the conjecture must use the lemma given in the input, we also re-weight the conjecturing examples at every iteration. Intuitively, if there is a distance function that can separate statements of different topics, the Wasserstein projection of the conjectures will have a similar distribution of topics, and therefore alleviates the mode collapsing issue.

\paragraph{Cost function.} We compute the cost $d(x,y)$ of matching conjecture $x$ to a statement $y$ by the negative of the cosine similarity between their embeddings, and the embedding is computed by the last hidden layer of the current model averaged over the sequence dimension. Since our model is trained to generate proofs of conjectures and statements, we expect that statements with similar proof techniques tend to have similar embeddings, and therefore smaller cost for the matching.

\paragraph{Algorithm.}
On the high level, our method computes a re-weighting of the generated conjectures that minimizes its Wasserstein distance to the unproved statements in the given dataset. Abstractly speaking, let $\calX$ be the set of generated conjectures, and $Q$ the set of unproved statements. Let $d(x,y)$ be the distance between a conjecture $x$ and a statement $y$. Then, the optimization problem can be written as
\begin{align}
	\argmin_{P:P\text{ is a valid distribution, }\supp(P)\subseteq\calX} W(P,Q),
\end{align}
where $W(P,Q)$ is the Wasserstein distance between $P$ and $Q$ (with little abuse of notation, we use $Q$ to represent the uniform distribution over the unproved statements). The Wasserstein distance $W(P,Q)$ is defined by the following optimal transportation problem where $\mu$ is a matching between the distribution $P$ and $Q$:
\begin{align}
	W(P,Q)=\min_\mu\quad &\sum_{x\in\supp(P), y\in \supp(Q)} \mu(x,y)d(x,y)\\
	\text{s.t.\quad}& \sum_{y\in\supp(Q)}\mu(x,y)=P(x),\\
			   & \sum_{x\in\supp(P)}\mu(x,y)=Q(y),\\
			   &\mu(x,y)\ge 0,\quad \forall x,y.
\end{align} 

Combining the equations above, the re-weighting distribution $P$ can be computed by 
\begin{align}
	\argmin_{P:\supp(P)\subseteq\calX}\min_{\mu} \quad &\sum_{x\in\supp(P), y\in \supp(Q)} \mu(x,y)d(x,y)\\
	\text{s.t.\quad}& \sum_{y\in\supp(Q)}\mu(x,y)=P(x),\\
	& \sum_{x\in\supp(P)}\mu(x,y)=Q(y),\\
	&\mu(x,y)\ge 0,\quad \forall x,y,\\
	& P(x)\ge 0, \quad \forall x,\\
	& \sum_{x\in\calX} P(x)=1,
\end{align}
where the last two constraint ensures that $P$ is a valid distribution. Equivalently, we get the following program,
\begin{align}
	\argmin_{P:\supp(P)\subseteq\calX}\min_{\mu} \quad &\sum_{x\in\supp(P), y\in \supp(Q)} \mu(x,y)d(x,y)\\
	\text{s.t.\quad}& \sum_{x\in\supp(P)}\mu(x,y)=Q(y),\\
	& \sum_{x\in\calX,y\in \supp(Q)} \mu(x,y)=1,\\
	&\mu(x,y)\ge 0,\quad \forall x,y,\\
	&P(x)=\sum_{y\in\supp(Q)}\mu(x,y).\label{equ:px-constrant}
\end{align}
Since $Q(y)$ is given, we can optimize $\mu(x,y)$ for every fixed $y$ separately, and then compute the final $P(x)$ using Eq.~\eqref{equ:px-constrant}. As a result, the program above has a closed-form solution $\mu^\star(x,y)=Q(y)\ind{x=\argmin_{x'\in\calX}d(x',y)}$ and $P(x)=\sum_{y\in\supp(Q)}\mu^\star(x,y)$. In other words, the optimal matching $\mu(x,y)$ for any given $y$ is only supported at the $x$ that minimizes the distance $d(x,y).$ Therefore, the (theoretical) algorithm that computes the optimal re-weighting is given in Alg.~\ref{alg:matching-theory}. Note that the last line in Alg.~\ref{alg:matching-theory} is to make sure that the sum of the weights equals the number of generated conjectures (i.e., the sum of weights before re-weighting).

\begin{algorithm}[htp]
	\caption{Computing the optimal re-weighting (theory).}
	\label{alg:matching-theory}
	\begin{algorithmic}[1]
		\State {\bfseries Input:} generated conjectures $\calX=\{x_1,\cdots,x_n\}$ of size $n$, unproved statements $Q$ with size $m$, and a distance function $d(x,y).$
		\State Initialize the optimal re-weighting $P=[0,0,\cdots,0]$.
		\For{$y\in Q$}
		\State Compute $x^\star=\argmin_{x\in \calX}d(x,y).$
		\State $P(x^\star)\gets P(x^\star)+1/m.$
		\EndFor
		\State {\bfseries Return:} the optimal re-weighting is $[P(x_1) * n, P(x_2) * n, \cdots, P(x_n) * n].$
	\end{algorithmic}
\end{algorithm}

Our practical implementation is shown in Alg.~\ref{alg:matching}. In this implementation, we additionally requires that the weighting $P$ for every conjecture $x$ cannot be too big because otherwise it might cause instability of the LLM training with weighted cross entropy loss. We also allow unproved statements in $Q$ to have different matching weights --- an important statement can be matched to more than one conjecture (see Line 5-6 of Alg.~\ref{alg:matching}). In both the Isabelle and Lean experiments, the statements from LeanWorkbook have matching weight 1. The statements from miniF2F-valid and ProofNet-valid have matching weight 1 for the first 24 iterations in the Lean experiment, and 128 afterward.

\begin{algorithm}[htp]
	\caption{Computing the optimal re-weighting.}
	\label{alg:matching}
	\begin{algorithmic}[1]
		\State {\bfseries Input:} generated conjectures $\calX=\{x_1,\cdots,x_n\}$ of size $n$, unproved statements $Q$ with size $m$, and a distance function $d(x,y).$
		\State Initialize the optimal re-weighting $P=[0,0,\cdots,0]$.
		\State Initialize the masks $M(x)=1,\forall x\in \calX.$
		\For{$y\in Q$}
			\State Let $k$ be the matching weight of $y$.
			\State Let $x^1,\cdots,x^k$ be the $k$ conjectures with smallest value of $d(\cdot,y)M(\cdot).$
			\For{$i=1,\cdots,k$}
				\State $P(x^i)\gets P(x^i)+1/m.$
				\If{$P(x^i) * n > 3$}
					\State $M(x^i)\gets 0.$
				\EndIf
			\EndFor
		\EndFor
		\State {\bfseries Return:} the optimal re-weighting is $[P(x_1) * n, P(x_2) * n, \cdots, P(x_n) * n].$
	\end{algorithmic}
\end{algorithm}

\subsection{Implementation details for expert iteration.}\label{app:expit-details}
In this section, we describe two different implementations of expert iteration and compare their performance.

\paragraph{Vanilla expert iteration.} For vanilla expert iteration, we only sample proofs to the \emph{unproved} statements in the given dataset. The LLM training dataset consists of all the correct proofs generated in this and previous iterations, and in each iteration, the model is trained from the base model.

\paragraph{Optimized expert iteration.} The most significant issue of vanilla expert iteration is the limited correct proofs generated in each iteration. As a result, even though the model is re-trained at every iteration, the difference between two models in consecutive iterations are limited. 

Therefore, in our optimized implementation of expert iteration, we generate proofs to all statements in the given dataset, regardless of whether they are previously proved or not. Then, to construct the LLM training dataset, we randomly choose at most 16 proofs per statement (so that the model does not overfit to the easy problems with many correct proofs). Note that this implementation requires slightly more sample budget per iteration. However, since the pass rate on the given dataset is low (less than 30\% even for our best model), this difference is not significant.

\begin{figure}[tp]
	\centering
	\includegraphics[width=.393\linewidth]{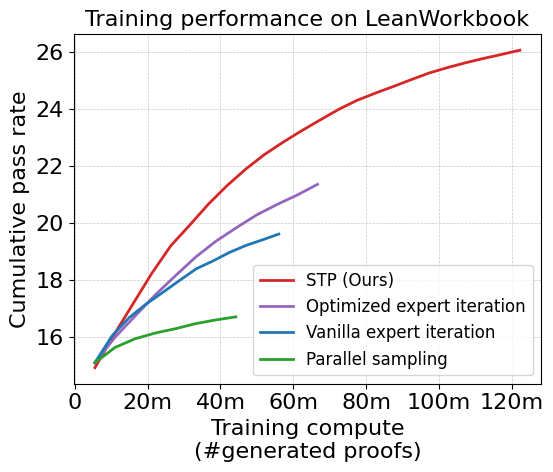}~~~~
	\includegraphics[width=.436\linewidth]{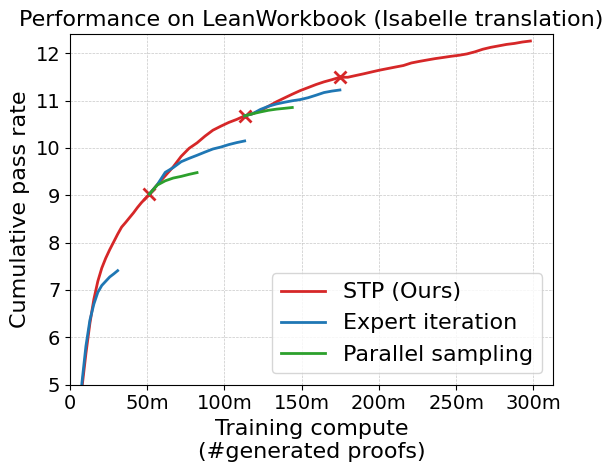}
	\caption{\textbf{Left:} Comparison of pass rates between \algname, two implementations of expert iteration, and parallel sampling methods on LeanWorkbook. \textbf{Right:} Comparison of pass rates between \algname and baseline methods on LeanWorkbook (Isabelle translation). The red crosses shows the points where we refresh the self-play training as described in Section~\ref{sec:implementation}.}
	\label{fig:expit-full}
\end{figure}

In Fig.~\ref{fig:expit-full} (Left), we plot the cumulative pass rate of two implementations of expert iteration, \algname and parallel sampling. \algname outperforms both implementations of expert iteration, and the optimized implementation of expert iteration is better than the vanilla implementation.

For the figures of Isabelle experiments, we always use the optimizes implementation of expert iteration. For Fig.~\ref{fig:lean-main}, we use the vanilla implementation.

\subsection{Additional details for interacting with the Isabelle verifier}\label{app:isabelle-detail}
For the Isabelle experiments, we have an additional filter for the conjectures --- if the generated conjecture is equivalent to the statement in the prompt (tested by \lstinline|solve_direct| in Isabelle), we consider it invalid.

We disallow the tactics \lstinline|sledgehammer, mason, smt, metis, sos| by invalidating proofs that contain any of these sub-strings. However, following the implementation of \citet{jiang2022thor}, we still use the keyword `sledgehammer' to replace the following simple tactics 
\begin{verbatim}
	[by auto, by simp, by blast, by fastforce, by force, by eval, by presburger, 
	by arith, by linarith, by (auto simp: field_simps)].
\end{verbatim}
During proof verification, we try these tactics sequentially to replace the keyword `sledgehammer'. If any of the tactics succeed, we proceed to the remaining proof steps. Otherwise we flag the proof incorrect.

\subsection{Additional details for interacting with the Lean4 verifier}\label{app:lean-detail}
During the self-play training, we use the same imports as the miniF2F Lean4 project\footnote{\url{https://github.com/yangky11/miniF2F-lean4/tree/main/MiniF2F}} instead importing the entire Mathlib to optimize the memory efficiency. This is because we do not have access to an additional CPU cluster for proof verification, and the available CPU memory in TPU-v4 VMs is limited.

\subsection{Compute resources}\label{app:compute}
Our experiments are primarily done on TPU-v4 VMs with 32 nodes. Each node contains 4 TPU chips (8 TPU cores), 240 CPU cores, and 400G memory. We use vLLM \citep{kwon2023efficient} to generate LLM outputs, and Levanter\footnote{\url{https://github.com/stanford-crfm/levanter}} to train the LLM. In both \algname and expert iteration, since the generated proofs are heavily filtered (based on the correctness, elegancy, trivialness, etc.) when constructing the training dataset, LLM training only takes less than 25\% of the wall-clock time for TPU compute, and generating proofs takes the rest 75\%.
\section{Additional Experiment Results}\label{app:additional-results}
In this section we show the additional experiment results with both Lean and Isabelle formal verifier.

\subsection{Additional results with Lean}\label{app:lean-results}
In Table~\ref{table:putnam}, we compare the performance of our method with prior works on PutnamBench. Note that DSP \citep{jiang2022draft} uses Isabelle verifier where PutnamBench only has 640 statements. Our model \algname achieves state-of-the-art performance by solving 8 out of 644 problems.

Table~\ref{table:stp-wo-conjectures} compares the model obtained by final re-training with and without the proofs of generated conjectures, as discussed in the ablation study section (Section~\ref{sec:ablation}). The results show that it is still beneficial to re-train with the generated conjectures in addition to the successfully proved statements in LeanWorkBook even for performance on miniF2F-test and ProofNet-test, which leads to about 2-3\% performance gain (for pass@128).

\begin{table}[!htb]
	\centering
	\captionof{table}{Pass rate on miniF2F and ProofNet.}
	\label{table:stp-wo-conjectures}
	\begin{small}
		\begin{tabular}{lccc}
			\toprule
			Method & Sample budget & MiniF2F-test & ProofNet-test \\ 
			\toprule 
			\text{\algname}~~\emph{(w/o conjectures)} & 128  & 58.3\% $\pm$ 0.7\% & 17.4\% $\pm$ 0.4\% \\
			\text{\algname} & 128 & 61.2\% $\pm$ 0.6\% & 19.5\% $\pm$ 0.7\% \\
			\bottomrule
		\end{tabular}
	\end{small}
\end{table}

\subsection{Additional results with Isabelle}\label{app:isabelle-results}
In Fig.~\ref{fig:expit-full} (Right), we plot the pass rates of \algname and baseline methods on LeanWorkbook starting from iteration 0. The red crosses shows the points where we refresh the training process as described in Section~\ref{sec:implementation}. Our models are tested with PutnamBench~\citep{tsoukalas2024putnambench}, commit \lstinline|d49896f|.\footnote{\url{https://github.com/trishullab/PutnamBench/tree/d49896fdc87a128a70e15a185d8dfca3516dd894}}

\begin{table*}[!htb]
	\caption{Results on PutnamBench.}
	\label{table:putnam}
	\begin{center}
		\begin{small}
			\begin{tabular}{lcc}
				\toprule
				Method & Sample budget (\#Proofs / \#Steps) & Result \\ 
				\toprule 
				\emph{Whole-Proof Generation Methods} \\\midrule
				DSP (GPT-4o) \citep{jiang2022draft} & 10 & 4/640 \\ 
				\algname & 128 & \textbf{7/644} \\
				& 3200 & \textbf{8/644} \\
				\midrule
				\emph{Tree Search Methods} \\\midrule
				InternLM2.5-StepProver-BF+CG & 2 $\times$ 32 $\times$ 600 &6/644 \\
				\bottomrule
			\end{tabular}
		\end{small}
	\end{center}
	\vskip -0.1in
\end{table*}

\subsection{Examples of unproved statements in LeanWorkbook}\label{app:leanworkbook-examples}

In this section, we list 20 randomly selected statements from LeanWorkbook that are not proved during STP training. The following table shows the formal statement, the corresponding natural language statement in LeanWorkbook, the correctness of formalization, and the correctness of the formal statement.

\begin{small}
\begin{xltabular}{\textwidth}{|c|X|X|p{1.7cm}|p{1.5cm}|}
\hline
	\# & Lean formal statement & Natural language statement & Correct & Correct \\
	&  &  & formalization? & statement? \\
\hline
1 &
\lstinline|theorem lean_workbook_7116 (x y z : ℝ) (hx : x + y + z = 3) : x ^ 2 + y ^ 2 + z ^ 2 + 3 ≤ 2 * (1 / x ^ 2 + 1 / y ^ 2 + 1 / z ^ 2)  :=  by| &
If $a=x^2, b=y^2, c=z^2$ it suffices to show that: $x+y+z=3\implies x^2+y^2+z^2+3\le 2(\frac{1}{x^2}+\frac{1}{y^2}+\frac{1}{z^2})$ & Yes & No. The case $x=0$ is ill defined. \\
\hline
2 &
\lstinline|theorem lean_workbook_plus_72390 (a b n : ℕ) (h : a ≡ b [ZMOD n]) : a^n ≡ b^n [ZMOD n^2]   :=  by| &
Prove that if $a \equiv b \pmod{n}$, then $a^n \equiv b^n \pmod{n^2}$. & Yes & Yes \\
\hline
3 &
\lstinline|theorem lean_workbook_35349 (a b c : ℝ) : (9 / (a + b + c + Real.sqrt (3 * (a * b + b * c + c * a)))) ≤ (1 / (a + b) + 1 / (b + c) + 1 / (c + a))  :=  by| &
For: $\frac{9}{a+b+c+\sqrt{3(ab+bc+ca)}}\leq\frac{1}{a+b}+\frac{1}{b+c}+\frac{1}{c+a}$ & Yes (but maybe missing the implicit assumption $a,b,c> 0$) & No (e.g., $(a,b,c)=(-0.5, 1, 1)$.)  \\
\hline
4 &
\lstinline|theorem lean_workbook_8880 (a b c : ℝ) : a * Real.sqrt (b ^ 2 + c ^ 2) + b * Real.sqrt (c ^ 2 + a ^ 2) + c * Real.sqrt (a ^ 2 + b ^ 2) ≤ 3 * Real.sqrt 2  :=  by| &
Prove that $ a\sqrt{b^2+c^2}+ b\sqrt{c^2+a^2}+ c\sqrt{a^2+b^2}\leq 3\sqrt{2},$ & Yes & No \\
\hline
5 &
\lstinline|theorem lean_workbook_plus_44018 (x : ℝ) (hx : 0 < x) (a : ℝ) (ha : a = x^(1/3)) : a^2 - 2*a - (a^3 - 4)*Real.sqrt (a^3 - 7) - 3*a^3 + 28 = 0   :=  by| &
Put $\sqrt[3] {x}=a$ . The equation is equivalent to $a^2-2a-(a^3-4)\sqrt{a^3-7}-3a^3+28=0$ & Yes & No \\
\hline
6 &
\lstinline|theorem lean_workbook_plus_35882 (a b c : ℝ) (ha : 0 < a) (hb : 0 < b) (hc : 0 < c) : (Real.sqrt ((a + 2 * b + 3 * c) / (4 * a + b + c)) + Real.sqrt ((3 * a + b + 2 * c) / (a + 4 * b + c)) + Real.sqrt ((2 * a + 3 * b + c) / (a + b + 4 * c))) ≥ 3   :=  by| &
If $a, b, c>0$ prove or disprove $\sqrt{\frac{a+2b+3c}{4a+b+c}}+\sqrt{\frac{3a+b+2c}{a+4b+c}}+\sqrt{\frac{2a+3b+c}{a+b+4c}}\geq3$ & Yes & Yes \\
\hline
7 &
\lstinline|theorem lean_workbook_12619 : ∀ x y : ℝ, (x^2+x+xy+y^2) ≤ 1 →  -(1/3)*Real.sqrt ((1/2)*(69+11*Real.sqrt 33)) ≤ x^2+2*x*y ∧ x^2+2*x*y ≤ (1/3)*Real.sqrt ((1/2)*(69+11*Real.sqrt 33))  :=  by| &
Let $x^2+x+xy+y^2\leq 1 (x, y \in R)$ . Prove that $-\frac{1}{3}\sqrt {\frac{1}{2}(69+11\sqrt{33})}\leq x^2+2xy\leq\frac{1}{3}\sqrt {\frac{1}{2}(69+11\sqrt{33})}$ & No (there is a xy term in Lean. Should be x * y) & No\\
\hline
8 &
\lstinline|theorem lean_workbook_plus_20629 (f : ℝ → ℝ) (x : ℝ) : f (f x + 1) = f x + 1   :=  by| &
Prove that $f(f(x)+1)=f(x)+1$ for all real $x$. & Yes & No \\
\hline
9 &
\lstinline|theorem lean_workbook_37208 (n : ℕ) (hn : 0 < n) : (n : ℝ) / (n! : ℝ) ^ (1 / n) < (1 + 1 / n)^n  :=  by| &
Prove that: ${ \frac{n}{\sqrt [n] {n!}}< \Big( 1+\frac{1}{n}\Big)^n} $ for every positive integer $n$ & Yes & Yes \\
\hline
10 &
\lstinline!theorem lean_workbook_10259 (a b : ℕ) (hab : a ≠ b) (h : a + b! $\mid$ \lstinline!a^2 + b^2) : a * b + 4 ≤ (Nat.gcd a b)^4  :=  by! &
Given $a,b \in \mathbb{N} (a \neq b)$ so that $a+b\mid a^2+b^2$ . Let $d=gcd(a,b)$ . Prove that $ab+4 \leq d^4$ & Yes & Yes \\
\hline
11 &
\lstinline|theorem lean_workbook_45322 (a b : ℝ) (ha : 0 < a) (hb : 0 < b) (hab : (a + 1 / a) * (b + 1 / b) = 2 + 3 / Real.sqrt 2) : 1 ≤ a ^ 4 + b ^ 4 ∧ a ^ 4 + b ^ 4 ≤ 4  :=  by| &
Let $a,b>0 $ and $ (a+\frac{1}{a})(b+\frac{1}{b})=2+\frac{3}{\sqrt 2} .$ Prove that $ 1\leq a^4+b^4\leq 4$ & Yes & Yes \\
\hline
12 &
\lstinline|theorem lean_workbook_28189 (x y z : ℝ) : Real.sqrt (1 + 48 * x / (y + z)) ≥ (184 * x ^ 2 - 32 * (y ^ 2 + z ^ 2) + 289 * x * (y + z) + 127 * y * z) / (8 * (x ^ 2 + y ^ 2 + z ^ 2) + 47 * (y * z + z * x + x * y))  :=  by| &
prove: $\sqrt{1+\frac{48x}{y+z}}\geq\frac{184x^{2}-32(y^{2}+z^{2})+289x(y+z)+127yz}{8(x^{2}+y^{2}+z^{2})+47(yz+zx+xy)}$ & Yes (but maybe missing $x\ge 0$) & No $x = -1/24, y = z = 1$ \\
\hline
13 &
\lstinline|theorem lean_workbook_31673  (x : ℝ)   (h₀ : ∑' k : ℕ, (7 / (2^k)) = x) :   x = 14  :=  by| &
Solution without Geometric Formula$\frac{7}{1} + \frac{7}{2} + \frac{7}{4} + \frac{7}{8}\cdots= x $ We divide everything by 2:  $\frac{7}{2} + \frac{7}{4} + \frac{7}{8} + \frac{7}{16}\cdots= \frac{x}{2} $ We substitute the original equation in:  $x-7=\frac{x}{2}$ $\frac{x}{2}=7$ Therefore, $\boxed{x=14}$ . & Yes & Yes \\
\hline
14 &
\lstinline|theorem lean_workbook_4086 (g : ℕ → ℕ) (h₁ : g 1 = g 1 ^ 2) : g 1 = 1  :=  by| &
Given $ g(1)=g(1)^2\Rightarrow g(1)=1$ & No (natural language statement is unclear) & No \\
\hline
15 &
\lstinline|theorem lean_workbook_6922 (a b : ℝ) (ha : 0 ≤ a) (hb : 0 ≤ b) (hab : 2 ≤ a + b) : a * Real.sqrt (a / (2 + 7 * b)) + b * Real.sqrt (b / (2 + 7 * a)) + Real.sqrt (1 / (1 + 8 * a * b)) ≥ 1  :=  by| &
Let $a,b \ge 0$ and $a+b\geq 2.$ Prove that $a\sqrt{\frac{a}{2+7b}}+b\sqrt{\frac{b}{2+7a}}+\sqrt{\frac{1}{1+8ab}} \geq 1$ & Yes & Yes \\
\hline
16 &
\lstinline|theorem lean_workbook_plus_65183 (f : ℝ → ℝ): (∀ x y, f (x + f y) = y + f (x + 1)) ↔ (∀ x, f x = x + 1) ∨ (∀ x, f x = -x + 1)   :=  by| &
Find all functions $f: \mathbb R \to \mathbb R$ such that $ f(x+f(y))=y+f(x+1),$ for all $x,y \in \mathbb R$. & Yes & No \\
\hline
17 &
\lstinline|theorem lean_workbook_26304 (a b c : ℝ) : a + b + c ≤ (a^2 * b^2 + b^2 * c^2 + c^2 * a^2) / (a * b * c)  :=  by| &
$ \Leftrightarrow a + b + c\le \frac {a^2b^2 + b^2c^2 + c^2a^2}{abc}$ & No (natural language statement is unclear) & No (e.g., $abc<0$ and $a+b+c>0$) \\
\hline
18 &
\lstinline|theorem lean_workbook_plus_30866 (x y z : ℝ) (hx : x^3 + y + z = 1) (hy : x + y^3 + z = 1) (hz : x + y + z^3 = 1) : x = y ∧ y = z ∧ z = x   :=  by| &
Solve the following system of equations:  $\begin{cases}x^3+y+z=1\\x+y^3+z=1\\x+y+z^3=1\end{cases}$ & No & No \\
\hline
19 &
\lstinline|theorem lean_workbook_plus_51637 (A : Matrix (Fin n) (Fin n) ℂ) (h : A * A.transpose = 0) : A = 0   :=  by| &
Let $A \in M_n ( \mathbb{C} )$ be so that $A \cdot A^t = O_n$ . Prove that $A=O_n$ . Here, $A^t$ is the transpose of $A$ . & Yes & No (not true for complex matrix) \\
\hline
20 &
\lstinline|theorem lean_workbook_plus_58359 (x y z : ℝ) (hx : 0 < x) (hy : 0 < y) (hz : 0 < z) : 1 ≤ x / (Real.sqrt (y * z)) * (1 / (x + 1)) + y / (Real.sqrt (z * x)) * (1 / (y + 1)) + z / (Real.sqrt (x * y)) * (1 / (z + 1)) ∧ x / (Real.sqrt (y * z)) * (1 / (x + 1)) + y / (Real.sqrt (z * x)) * (1 / (y + 1)) + z / (Real.sqrt (x * y)) * (1 / (z + 1)) ≤ Real.sqrt 2   :=  by| &
We also have a nice inequality  $ 1 \leq \dfrac{x}{\sqrt{yz}}\cdot \dfrac{1}{x+1}+\dfrac{y}{\sqrt{zx}}\cdot \dfrac{1}{y+1}+\dfrac{z}{\sqrt{xy}}\cdot \dfrac{1}{z+1} \leq \sqrt{2}.$  With $ 1 $ and $ \sqrt{2} $ are the best constant. & Yes & No (when $x, y, z = \epsilon\to 0$, this term goes to 3) \\
	\hline
\end{xltabular}
\end{small}

\end{document}